\documentclass{sip}%%%%where SIP is the template name

\usepackage{amsmath} 
\usepackage{amsthm} 
\usepackage{hyperref} 
\usepackage{graphics}
\usepackage{algorithmic} 
\usepackage{subfig} 
\usepackage{url}
\usepackage{lipsum}
\usepackage{mathtools, cuted}

\usepackage{epstopdf, epsfig}

\begin{document}

\title[Towards Visible and Thermal Drone Monitoring with Convolutional Neural Networks]{Towards Visible and Thermal Drone Monitoring with Convolutional Neural Networks}

\author[Ye Wang, \textit{et al}.]{Ye Wang$^{1}$, Yueru Chen$^{1}$, Jongmoo Choi$^{1}$ and C.-C. Jay Kuo$^{1}$}

\address{\add{1}{University of Southern California, Los Angeles, CA 90089, USA}}
% \add{*}{indicates equal contributions}}

\corres{\name{C.-C. Jay Kuo}
\email{cckuo@sipi.usc.edu}}

\begin{abstract}
This paper reports a visible and thermal drone monitoring system that integrates deep-learning-based detection and tracking modules.
The biggest challenge in adopting deep learning methods for drone detection is the paucity of training drone images especially thermal drone images. To address this issue, we develop two data augmentation techniques. One is a
model-based drone augmentation technique that automatically generates visible
drone images with a bounding box label on the drone's location. The other is exploiting an adversarial data augmentation methodology to create thermal drone images. To track a
small flying drone, we utilize the residual information between
consecutive image frames. Finally, we present an integrated detection
and tracking system that outperforms the performance of each individual
module containing detection or tracking only. The experiments show that,
even being trained on synthetic data, the proposed system performs well
on real world drone images with complex background. The USC drone
detection and tracking dataset with user labeled bounding boxes is
available to the public. 
\end{abstract}


\maketitle

\section{Introduction}
There is a growing interest in the commercial and recreational use of
drones. This in turn imposes a threat to public safety.  The Federal
Aviation Administration (FAA) and NASA have reported numerous cases of
drones disturbing the airline flight operations, leading to near
collisions. It is therefore important to develop a robust drone
monitoring system that can identify and track illegal drones.  Drone
monitoring is however a difficult task because of diversified and
complex background in the real world environment and numerous drone
types in the market.  

%%%%%%%%%%%%%%%%%%%%%%%%%%%%%%%%%%%%%
\begin{figure*}[t]
\begin{center}
\includegraphics[width=0.95\linewidth]{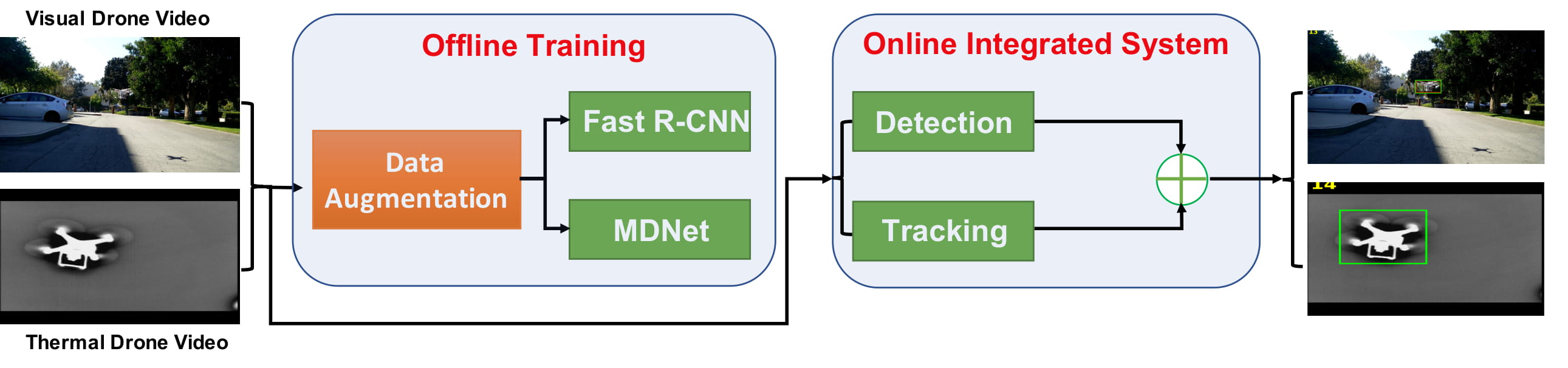}
\end{center}
\caption{Overview of proposed approach. We integrate the tracking module and detector module to set up an integrated system. The integrated system can monitor the drone during day and night with exploiting our proposed data augmentation techniques.}\label{fig:overview_all}
\end{figure*}
%%%%%%%%%%%%%%%%%%%%%%%%%%%%%%%%%%%%%

Generally speaking, techniques for localizing drones can be categorized
into two types: acoustic and optical sensing techniques.  The acoustic
sensing approach achieves target localization and recognition by using a
miniature acoustic array system.  The optical sensing approach processes
images or videos to estimate the position and identity of a target
object. In this work, we employ the optical sensing approach by
leveraging the recent breakthrough in the computer vision field. 

The objective of video-based object detection and tracking is to detect
and track instances of a target object from image sequences. In earlier
days, this task was accomplished by extracting discriminant features
such as the scale-invariant feature transform (SIFT) \cite{sift} and the
histograms of oriented gradients (HOG) \cite{hog}.  The SIFT feature
vector is attractive since it is invariant to object's translation,
orientation and uniform scaling. Besides, it is not too sensitive to
projective distortions and illumination changes since one can transform an
image into a large collection of local feature vectors. The HOG feature
vector is obtained by computing normalized local histograms of image
gradient directions or edge orientations in a dense grid. It provides
another powerful feature set for object recognition. 

In 2012, Krizhevsky {\em et al.} \cite{alexnet} demonstrated the power
of the convolutional neural network (CNN) in the ImageNet grand
challenge, which is a large scale object classification task,
successfully. This work has inspired a lot of follow-up work on the
developments and applications of deep learning methods. A CNN consists
of multiple convolutional and fully-connected layers, where each layer
is followed by a non-linear activation function. These networks can be
trained end-to-end by back-propagation.  There are several variants in
CNNs such as the R-CNN \cite{rcnn}, SPPNet \cite{sppnet} and Faster-RCNN
\cite{fasterrcnn}.  Since these networks can generate highly
discriminant features, they outperform traditional object detection
techniques by a large margin. The Faster-RCNN
includes a Region Proposal Network (RPN) to find object proposals, and
it can reach nearly real-time object detection. 

In particular, our proposed model integrates the detector module and tracker module to set up a drone monitoring system as illustrated in Fig. \ref{fig:overview_all}. The proposed system can monitor drones during both day and night. Due to the lack of the drone data and paucity of thermal drone diversities, we propose model-based augmentation for visible drone data augmentation and design a modified Cycle-GAN-based generation approach for thermal drone data augmentation. Furthermore, a residual tracker module is presented to deal with fast motion and occlusions. Finally, we demonstrate the effectiveness of the proposed integrated model on USC drone dataset and attain an AUC score of 43.8\% on the test set.

The contributions of our work are summarized below.
\begin{itemize}
\item To the best of our knowledge, this is the first one to use the
deep learning technology to solve the challenging drone detection and
tracking problem. 
\item We propose to exploit a large number of synthetic drone images, which
are generated by conventional image processing and 3D rendering algorithms,
along with a few real 2D and 3D data to train the CNN. 
\item We develop an adversarial data augmentation technique, a modified Cycle-GAN-based generation approach, to create more thermal drone images to train the thermal drone detector.
\item We propose to utilize the residue information from an image sequence
to train and test an CNN-based object tracker. It allows us to track a
small flying object in the cluttered environment. 
\item We present an integrated drone monitoring system that consists of a
drone detector and a generic object tracker. The integrated system
outperforms the detection-only and the tracking-only sub-systems. 
\item We have validated the proposed system on USC drone dataset.
\end{itemize}

The rest of this paper is organized as follows. Related work is reviewed in Sec. \ref{sec:related}. The collected drone
datasets are introduced in Sec. \ref{sec:dataset}. The proposed drone
detection and tracking system is described in Sec.  \ref{sec:solution}.
Experimental results are presented in Sec.  \ref{sec:results}.
Concluding remarks are given in Sec.  \ref{sec:conclusion}.

\section{Related Work}\label{sec:related} 

\subsection{Object Detection}

Current state-of-the-art CNN object detection approaches include two main streams: two-step and one-step object detection. 

Two-step object detection approaches are based on R-CNN \cite{rcnn} framework, the first step generates candidate object bounding box and the second step classifies each candidate bounding box as foreground or background using a convolutional neural network. The R-CNN method \cite{rcnn} trains CNNs end-to-end to classify the proposal regions into object categories or background. SPPnet \cite{sppnet} develops spatial pyramid pooling on shared convolutional feature maps for efficient object detection and semantic segmentation. Inspired by SPPnet, Fast R-CNN \cite{fastrcnn} enables shared computation on the entire image and then the detector network evaluates the individual regions which dramatically improves the speed. Faster R-CNN \cite{fasterrcnn} proposes a Region Proposal Network (RPN) to generate candidate bounding boxes followed by a second-step classifier which is the same as that of Fast R-CNN. The two-step framework consistently achieves top accuracy on the challenging COCO benchmark \cite{lin2014microsoft}. 

One-step object detection approaches predict bounding boxes and confidence scores for multiple categories directly without the proposal generation step in order to reduce the training and testing time. OverFeat \cite{sermanet2013overfeat}, a deep multiscale and sliding window method, presents an integrated framework to implement classification, localization and detection simultaneously by utilizing a single shared network. YOLO \cite{redmon2016you} exploits the whole topmost feature map to predict bounding boxes and class probabilities directly from full images in one evaluation. SSD \cite{liu2016ssd} utilizes default boxes of different aspect ratios and scales on each feature map location. At prediction time, the classification scores are determined in each default box and the bounding box coordinates are adjusted to match the object shape. To handle objects of various sizes, multiple feature maps with different resolutions are combined to perform better predictions. 

\subsection{Object Tracking}

Object tracking is one of the fundamental problems in computer vision and CNN-based tracking algorithms have developed very fast due to the development of deep learning. The trackers can be divided to three main streams: correlation filter-based trackers \cite{galoogahi2017learning,danelljan2017eco}, Siamese network-based trackers \cite{tao2016siamese,bertinetto2016fully} and detection based trackers \cite{mdnet,jung2018real}. 

Correlation filter-based tracker can detect objects very fast in the frequency domain. Recent techniques incorporate representations from convolutional neural networks with discriminative correlation filters to improve the performance. BACF \cite{galoogahi2017learning} designs a new correlation filter by learning from negative examples densely extracted from background region. ECO \cite{danelljan2017eco} proposes an efficient discriminative correlation filter for visual tracking by reducing the number of parameters and designing a compact generative model.  

The Siamese network is composed with two-branch CNNs with tied parameters, and takes the image pairs as input and predict their similarities. Siamese network based trackers learn a matching function offline on image pairs. In the online tracking step, the matching function is exploited to find the most similar object region compared with the object in the first frame. SiamFC \cite {bertinetto2016fully} trains a fully covolutional Siamese network directly without online adaptation, and it achieves 86 fps with GPU but its tracking accuracy is not state-of-the-art. SINT \cite{tao2016siamese} utilizes optical flow to deal with candidate sampling and it achieves higher tracking accuracy but lower speed. CFNet \cite{valmadre2017end} interprets the correlation filter as a differentiable layer in a deep neural network to compute the similarity between the two input patches. The experimental results show comparable tracking accuracy at high framerates. 

Tracking-by-detection approaches train a classifier to distinguish positive image patches with negative image patches. MDNet \cite{mdnet} finetunes a classification network to learn class-agnostic representations appropriate for visual tracking task. It proposes a multi-domain learning framework to separate the domain-independent information from the domain-specific one. Although MDNet demonstrates state-of-the-art tracking accuracies on two benchmark datasets, the tracking speed is about 1 fps. RT-MDNet \cite{jung2018real} utilizes improved RoIAlign \cite{he2017mask} technique to improve the tracking speed by extracting representations from the feature map instead of the image. This approach achieves similar accuracy with MDNet with real-time tracking speed.

\subsection{Generative Adversarial Network}

Paucity of thermal drone training data forms a major bottleneck in training deep neural networks for drone monitoring. To address the problem, we propose an adversarial data generation approach to augment the existing thermal drone data. 

Generative Adversarial Networks (GANs) simultaneously train two models: a generator and a discriminator. The generator tries to generate data from some distributions to maximize the probability of the discriminator making a mistake, while the discriminator distinguishes the sample comes from the training data rather than the generator. GANs have showed impressive results in a wide range of tasks, such as generating high-quality images \cite{radford2015unsupervised}, semi-supervised learning \cite{salimans2016improved}, image inpainting \cite{yeh2016semantic}, video prediction and generation \cite{mathieu2015deep}, and image translation \cite{isola2017image}. Current image-to-image translation approaches have drawn more and more attentions due to the development of GANs. Pix2Pix \cite{isola2017image} exploits a regression loss to guide the GAN to learn pairwise image-to-image translation. Due to the lack of the paired data, Cycle-GAN \cite{zhu2017unpaired} utilizes a combination of adversarial and cycle-consistent losses to deal with unpaired data image-to-image translation. Taigman \textit{et al}. \cite{taigman2016unsupervised} exploits cycle-consistency in the feature map with the adversarial loss to transfer a sample in one domain to an analog sample in another domain. In our paper, a new unpaired image-to-image translation algorithm is proposed to augment the thermal drone images.

\section{Data Collection and Augmentation}\label{sec:dataset} 

%%%%%%%%%%%%%%%%%%%%%%%%%%%%%%%%%%%%%
\begin{figure*}[ht]
\centering 
\subfloat[Public-Domain Drone Dataset]{\includegraphics[width=0.45\linewidth]{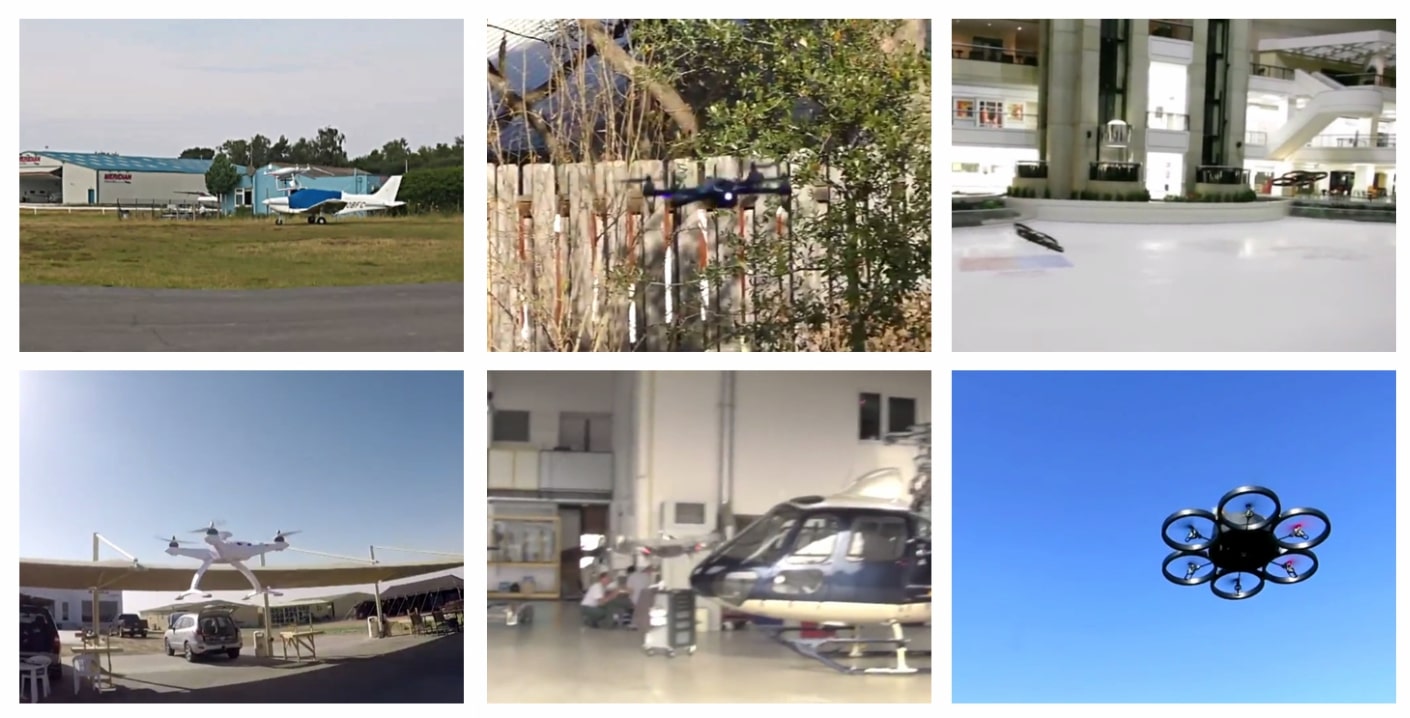}\label{fig:dataset2}}  \hfil 
\subfloat[USC Drone Dataset]{\includegraphics[width=0.45\linewidth]{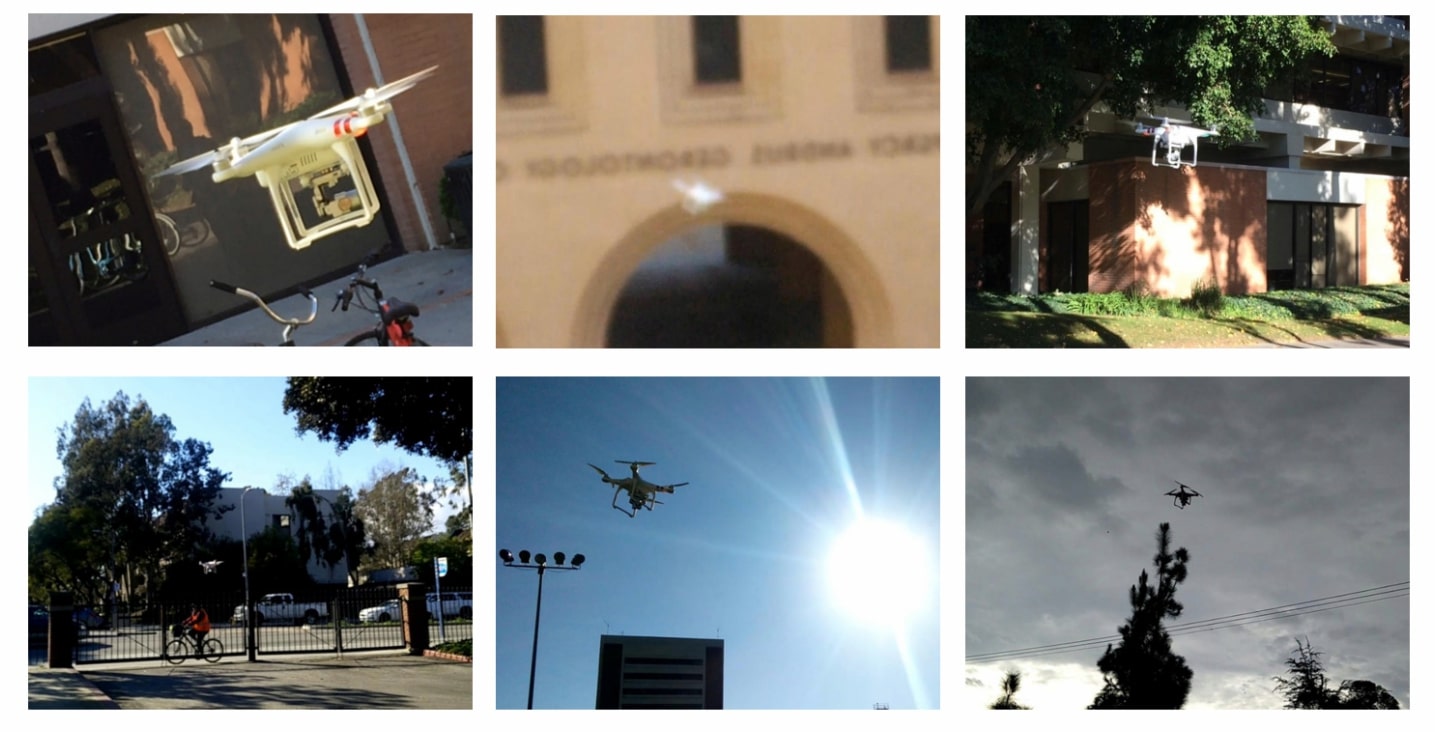} \label{fig:dataset1}} \hfil 
\caption{Sampled frames from two collected drone datasets.}\label{fig:dataset}
\end{figure*}
%%%%%%%%%%%%%%%%%%%%%%%%%%%%%%%%%%%%%

\subsection{Data Collection}

The first step in developing the drone monitoring system is to collect
drone flying images and videos for the purpose of training and testing.
We collect two drone datasets as shown in Fig. \ref{fig:dataset}. They
are explained below.
\begin{itemize}
\item Public-Domain drone dataset. \\
It consists of 30 YouTube video sequences captured in an indoor or outdoor
environment with different drone models.  Some samples in this dataset are
shown in Fig.  \ref{fig:dataset2}. These video clips have a frame
resolution of 1280 x 720 and their duration is about one minute.  Some
video clips contain more than one drone. Furthermore, some shoots are
not continuous. 
\item USC drone dataset. \\
% thermal drone dataset
It contains 30 visible video clips shot at the USC campus. All of them were shot with a single drone model. Several examples of the same drone in
different appearance are shown in Fig. \ref{fig:dataset1}. To shoot
these video clips, we consider a wide range of background scenes,
shooting camera angles, different drone shapes and weather conditions.
They are designed to capture drone's attributes in the real world such
as fast motion, extreme illumination, occlusion, etc.  The duration of
each video is approximately one minute and the frame resolution is 1920 x
1080. The frame rate is 30 frames per second. 
\item USC thermal drone dataset. \\
It contains 10 thermal video clips shot at the USC campus and sample images are demonstrated in Fig. \ref{fig:thermal_dataset}. All of them were shot with the same drone model as that for USC drone dataset.  
Each video clip is approximately one minute and the frame resolution is 1920 x 1080. The frame rate is 30 frames per second. 
\end{itemize}

We annotate each drone sequence with a tight bounding box around the
drone. The ground truth can be used in CNN training. It can also be used
to check the CNN performance when we apply it to the testing data.

\subsection{Data Augmentation}\label{sec:augmentation}

The preparation of a wide variety of training data is one of the main
challenges in the CNN-based solution.  For the drone monitoring task,
the number of static drone images is very limited and the labeling of
drone locations is a labor intensive job. The latter also suffers from
human errors. All of these factors impose an additional barrier in
developing a robust CNN-based drone monitoring system. To address this
difficulty, we develop a model-based data augmentation technique that
generates training images and annotates the drone location at each frame
automatically. 

\begin{figure}[t]
\begin{center}
\includegraphics[width=0.95\linewidth]{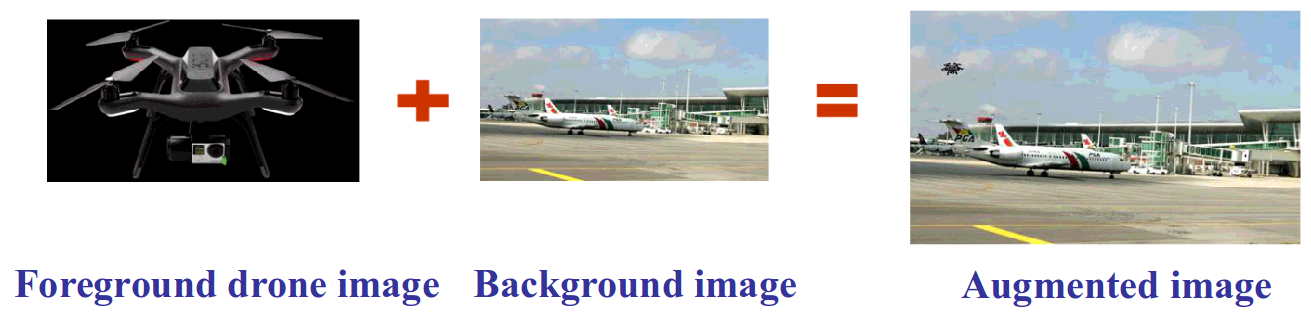}
\end{center}
\caption{Illustration of the data augmentation idea, where augmented
training images can be generated by merging foreground drone images and
background images.}\label{fig:augment}
\end{figure}

\begin{figure}[ht]
\begin{center}
\includegraphics[width=0.95\linewidth]{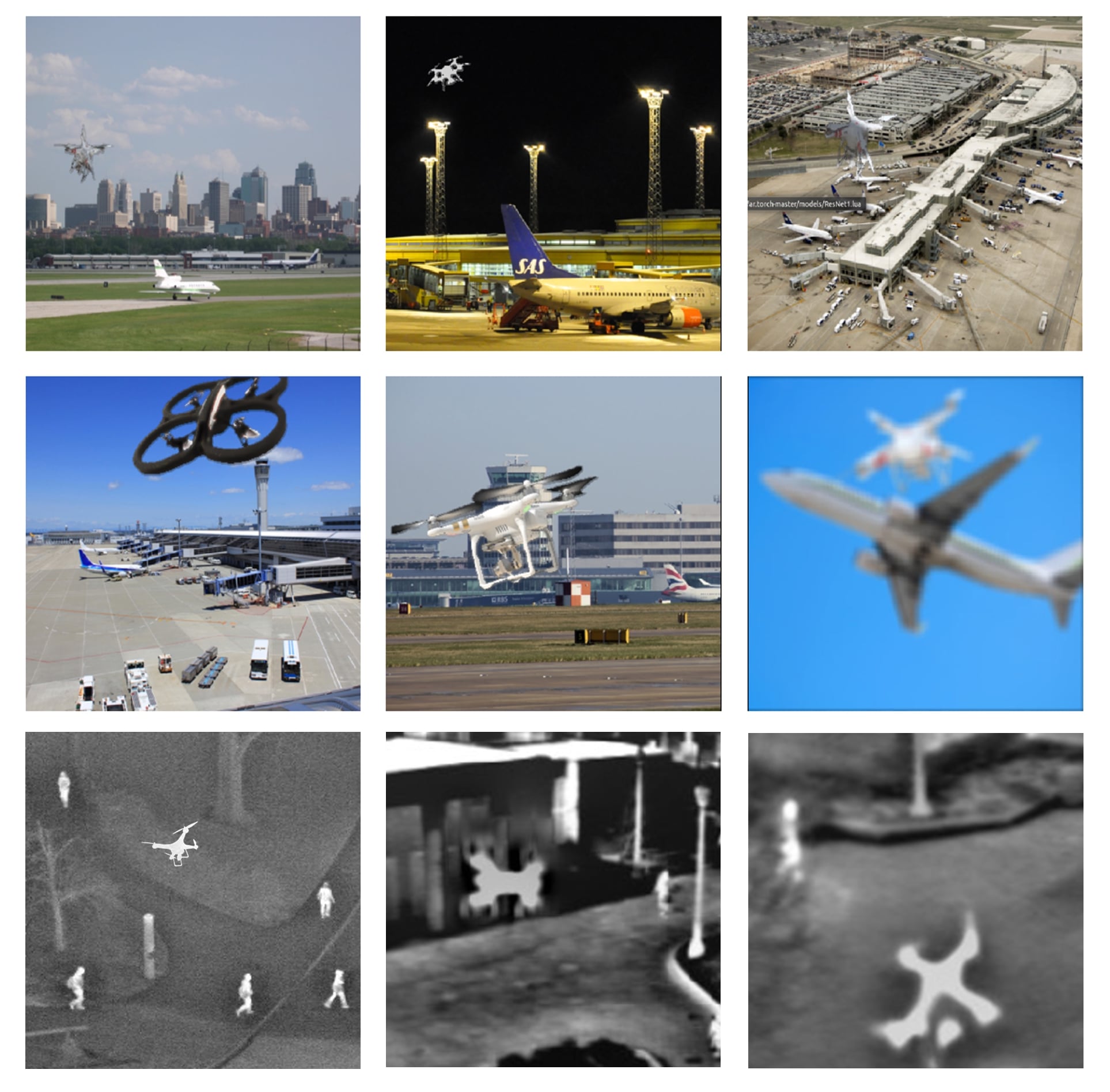}
\end{center}
\caption{Synthesized visible and thermal images by incorporating various illumination conditions, image qualities,
and complex backgrounds.}\label{fig:augresult}
\end{figure}

The basic idea is to cut foreground drone images and paste them on top
of background images as shown in Fig. \ref{fig:augment}. To accommodate
the background complexity, we select related classes such as aircrafts and cars in the PASCAL VOC 2012 \cite{pascal}. As to the diversity of drone
models, we collect 2D drone images and 3D drone meshes of many drone
models. For the 3D drone meshes, we can render their corresponding
images by changing the view-distance, viewing-angle, and lighting
conditions of the camera. As a result, we can generate many different drone images
flexibly.  Our goal is to generate a large number of augmented images to
simulate the complexity of background images and foreground drone models
in a real world environment.  Some examples of the augmented drone images of
various appearances are shown in Fig. \ref{fig:augresult}. 

%%%%%%%%%%%%%%%%%%%%%%%%%%%%%%%%%%%%%
%%%%%%%%%%%%%%%%%%%%%%%%%%%%%%%%%%%%%

Specific drone augmentation techniques are described below.
\begin{itemize} 
\item Geometric transformations \\
We apply geometric transformations such as image translation, rotation
and scaling. We randomly set the width of the foreground drone in the range (0.1, 0.5) of the background image width, and keep the height-width ratio unaltered. We randomly select the angle of rotation from the range
(-30$^{\circ}$, 30$^{\circ}$). Furthermore, we conduct uniform scaling
on the original foreground drone images along the horizontal and the
vertical direction. Finally, we randomly select the drone location in
the background image. 
\item Illumination variation \\
To simulate drones in the shadows, we generate regular shadow maps by
using random lines and irregular shadow maps via Perlin noise
\cite{perlin}. In the extreme lighting environments, we observe that
drones tend to be in monochrome (i.e. the gray-scale) so that we 
change drone images to gray level ones. 
\item Image quality \\
This augmentation technique is used to simulate blurred drones caused
by camera's motion and out-of-focus. We use some blur filters (e.g.  the
Gaussian filter, the motion Blur filter) to create the blur effects on
foreground drone images. 
\end{itemize}

%%%%%%%%%%%%%%%%%%%%%%%%%%%%%%%%%%%%%
\begin{figure}[ht]
\begin{center}
\includegraphics[width=0.95\linewidth]{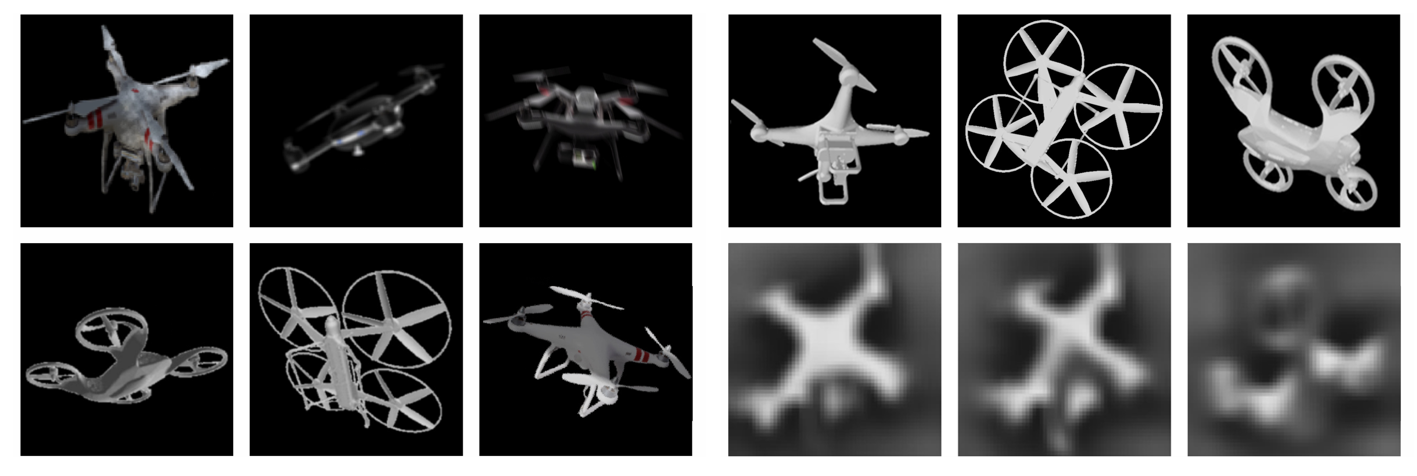}
\end{center}
\caption{Illustration of augmented visible and thermal drone models. The left three columns show the augmented visible drone models using different augmentation techniques. The right three columns show the augmented thermal drone models with the first row exploiting 3D rendering technique and the second row utilizing Generative Adversarial Networks. }\label{fig:augresultboth}
\end{figure}
%%%%%%%%%%%%%%%%%%%%%%%%%%%%%%%%%%%%%

%%%%%%%%%%%%%%%%%%%%%%%%%%%%%%%%%%%%%
%%%%%%%%%%%%%%%%%%%%%%%%%%%%%%%%%%%%%

%%%%%%%%%%%%%%%%%%%%%%%%%%%%%%%%%%%%%
\begin{figure}[ht]
\begin{center}
\includegraphics[width=0.95\linewidth]{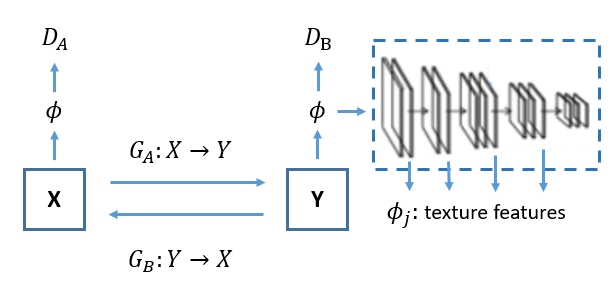}
\end{center}
\caption{The architecture of the proposed thermal drone generator.}\label{fig:cycle}
\end{figure}
%%%%%%%%%%%%%%%%%%%%%%%%%%%%%%%%%%%%%
We use the model-based augmentation technique to acquire more training images with the ground-truth labels and show several exemplary synthesized drone images in Fig. \ref{fig:augresult}, where augmented drone models are shown in Fig. \ref{fig:augresultboth}. 

\subsection{Thermal Data Augmentation}\label{sec:th-augmentation}
In real-life applications, our systems are required to work in both daytime and nighttime. To monitor the drones efficiently during the nighttime, we train our CNN-based thermal drone detector using infrared thermal images. It is more difficult to acquire enough training data for training the thermal drone detector. We can therefore apply data augmentation methods as mentioned in the previous section to generate thermal drone images with drone bounding box annotations. As illustrated in Fig. \ref{fig:augment}, we collect thermal images as the background from public thermal datasets and the Internet.

However, it is difficult to directly apply the visible data augmentation techniques since the thermal drone models are very limited and we cannot collect enough foreground thermal drone models with large diversity. This problem can be solved if we can successfully translate a visible drone image to a corresponding thermal drone image, where we face an unsupervised image-to-image translation problem. To address this issue, we provide two approaches for generating thermal foreground drone images. One is specifically targeting at translating thermal drone images using traditional image processing techniques, and the other is the proposed image translation approach using GANs. 

\begin{itemize}
\item Traditional image processing techniques 

From observation of USC thermal drone dataset in Fig. \ref{fig:thermal_dataset}, thermal drones have nearly uniform gray color in most cases. Therefore a post-processing section is added to convert visible drones to monochrome drones. 

%%%%%%%%%%%%%%%%%%%%%%%%%%%%%%%%%%%%%
\begin{figure}[t]
\begin{center}
\includegraphics[width=0.95\linewidth]{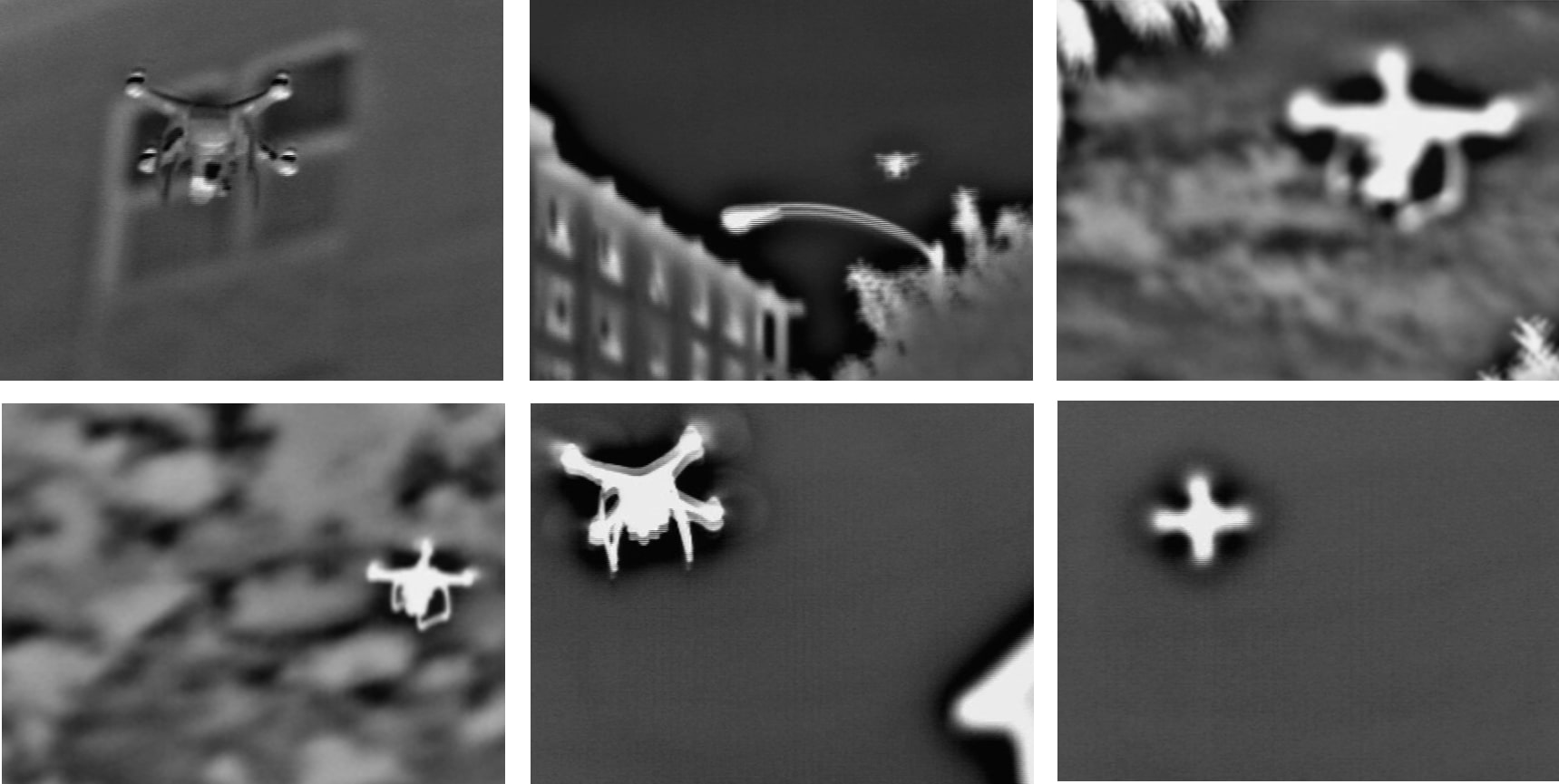}
\end{center}
\caption{Sampled frames from collected USC thermal drone dataset.}\label{fig:thermal_dataset}
\end{figure}
%%%%%%%%%%%%%%%%%%%%%%%%%%%%%%%%%%%%%

\item Modified Cycle-GAN \\
Our goal is to learn mapping functions between two domains X and Y given the unbalanced training samples. In our case, there are enough samples in visible domain X but very few samples in thermal domain Y, and applying the learned mapping function helps to generate a large diversity of samples in domain Y. 

Cycle-GAN provides a good baseline for unpaired image-to-image translation problem, however the training images in two domains are heavily imbalanced in our case. As demonstrated in the second row in Fig. \ref{fig:thermal_compare}, Cycle-GAN cannot increase the diversity of drone foreground images. Their proposed cycle consistency loss is, however, necessary to solve our problem. We utilize this cycle consistency loss to constrain the object shape consistency in two domains. The objective also contains the perceptual texture loss for only learning the texture translation between two domains, which helps to address the failures of Cycle-GAN. 

%%%%%%%%%%%%%%%%%%%%%%%%%%%%%%%%%%%%%
\begin{figure}[t]
\begin{center}
\includegraphics[width=0.95\linewidth]{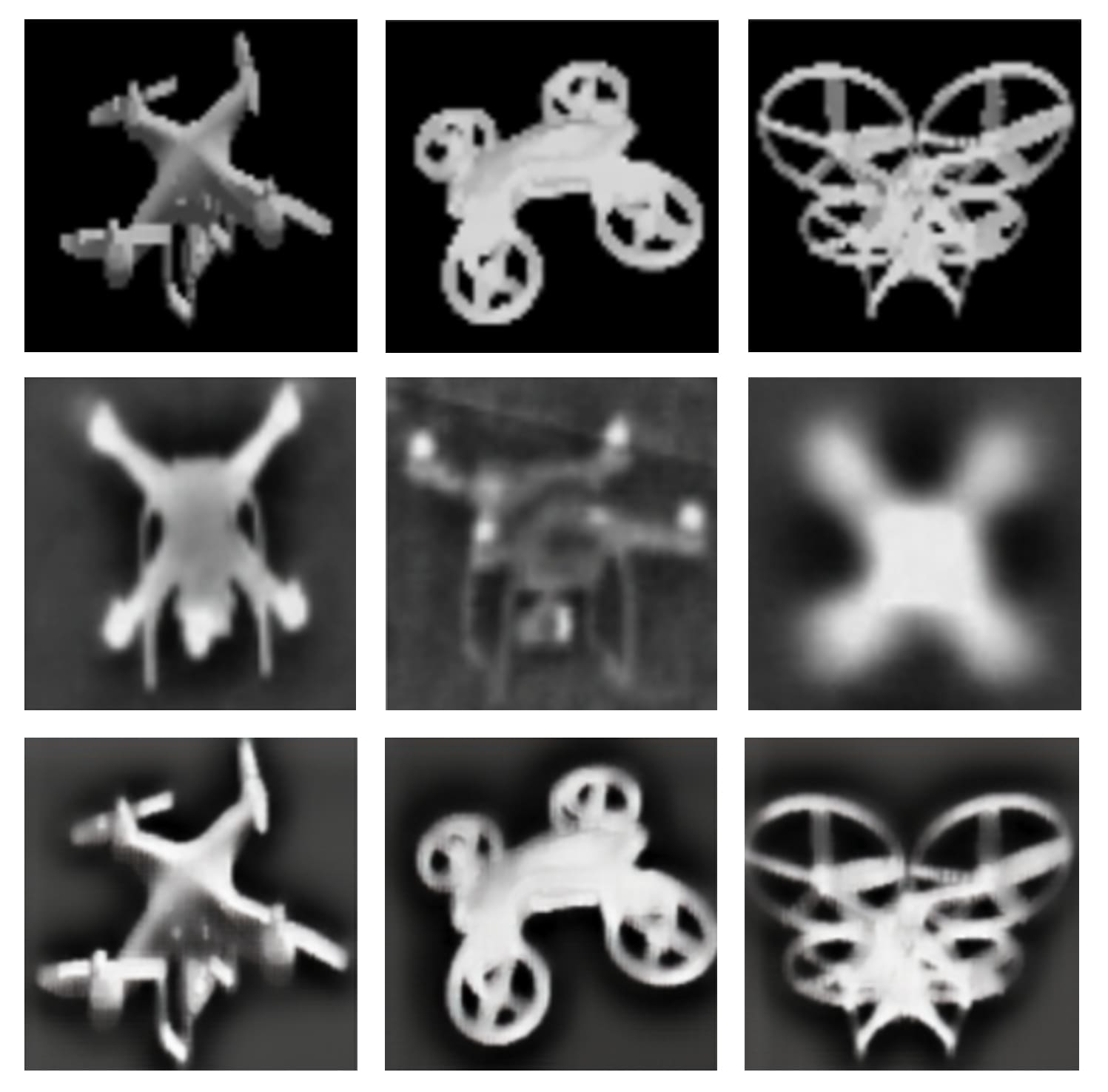}
\end{center}
\caption{Comparison of generated thermal drone images of different methods: 3D rendering (first row), Cycle-GAN (second row), proposed method (third row). }\label{fig:thermal_compare}
\end{figure}
%%%%%%%%%%%%%%%%%%%%%%%%%%%%%%%%%%%%%

As illustrated in Fig. \ref{fig:cycle}, our methods learn two generators $G_A : X \rightarrow Y$ and $G_B : Y \rightarrow X$ with corresponding discriminator $D_A$ and $D_B$. Image $x \in X$ is translated to domain $Y$ as $G_A(x)$ and translated back as $G_B(G_A(x))$ which should be the reconstruction of $x$. Similarly, image $y \in Y$ is translated to domain $X$ as $G_B(y)$ and translated back as $G_A(G_B(y))$. The cycle consistency loss is defined as the sum of the two reconstruction loss: 

% \lipsum[1]
% \begin{equation} \label{eqn:confidence}
% \resizebox{0.42\textwidth}{!}{$L_{cycle}(G_A, G_B, X, Y) = \| G_B(G_A(x)) - x\| \\ + \| G_A(G_B(y)) - y\|$}
% \end{equation}  

% \lipsum[1]
\begin{eqnarray} \label{eqn:confidence}
\nonumber
&L_{cycle}(G_A, G_B, X, Y) = E_{x\sim P_x}[\| G_B(G_A(x)) - x\|] \\
&+ E_{y\sim P_y}[\| G_A(G_B(y)) - y\|]
\end{eqnarray}

 We extract texture features of images as inputs of discriminators which aims to distinguish the texture styles between images $x$ and translated images $G_B(y)$, images $y$ and translated images $G_A(x)$. We exploit the texture features proposed by Gatys \textit{et al} \cite{gatys2016image}, which is the gram matrix $G$ of network feature map of layer $j$. The perceptual texture GAN loss is defined as:

\begin{eqnarray} \label{eqn:confidence}
\nonumber
&L_{tex}(G_A, D_B , X, Y) = E_{y\sim P_y}[\log D_B (G(\phi_j (y)))] \\ 
&+ E_{x\sim P_x}[\log (1 - D_B (G_A(G(\phi_j (x)))))]
\end{eqnarray}

% \begin{eqnarray} \label{eqn:confidence}
% \nonumber
% L_{gan-tex}(G_A, D_B , X, Y) &=& \log [1 - D_B [G_A(G(\phi_j (x)))]] \\ 
% &+& [t\log D_B (G(\phi_j (y)))]
% \end{eqnarray}

The full loss function is:

\begin{eqnarray} \label{eqn:confidence}
 % \small
 % \begin{aligned}[b]
 \nonumber
&L_{loss}(G_A, D_A , G_B, D_B) = \lambda L_{cycle}(G_A, G_B, X, Y) \\
&+ L_{tex}(G_A, D_B , X, Y) + L_{tex}(G_B, D_A , Y, X)
 % \end{aligned}
 \end{eqnarray}

% \normalsize
% \scriptsize
% In the experiments, we use VGG network pretrained on ImageNet.
\end{itemize}

\noindent where $\lambda$ controls the relative importance of the cycle consistency loss and perceptual texture GAN loss.

\section{Drone Monitoring System}\label{sec:solution}

To achieve the high performance, the system consists of two modules;
namely, the drone detection module and the drone tracking module. Both
of them are built with the deep learning technology. These two modules
complement each other, and they are used jointly to provide the accurate
drone locations for a given video input. 

\subsection{Drone Detection}\label{sec:detection}

The goal of drone detection is to detect and localize the drone in
static images. Our approach is built on the Faster-RCNN
\cite{fasterrcnn}, which is one of the state-of-the-art object detection
methods for real-time applications. The Faster-RCNN utilizes the deep
convolutional networks to efficiently classify object proposals. To
achieve real time detection, the Faster-RCNN replaces the usage of
external object proposals with the Region Proposal Networks (RPNs) that
share convolutional feature maps with the detection network. The RPN is
constructed on the top of convolutional layers. It consists of two
convolutional layers, one encodes conv feature maps for each
proposal to a lower-dimensional vector and the other provides the
classification scores and regressed bounds. The Faster-RCNN achieves
nearly cost-free region proposals and it can be trained end-to-end by
back-propagation.  We use the Faster-RCNN to build the drone detector by
training it with synthetic drone images generated by the proposed data
augmentation technique for the daytime case and by the proposed thermal data augmentation method for the nighttime case as described in Sec. \ref{sec:dataset}. 

\subsection{Drone Tracking}\label{sec:tracking}

The drone tracker attempts to locate the drone in the next frame based
on its location at the current frame. It searches around the
neighborhood of the current drone's position. This helps track a drone
in a certain region instead of the entire frame. To achieve this
objective, we use the state-of-the-art object tracker called the
Multi-Domain Network (MDNet) \cite{mdnet} as the backbone. Due to the fact that learning a unified representation across different video sequences is challenging and the same object class can be considered not only as a foreground but also background object. The MDNet is able to separate the domain specific information from the domain independent
information in network training. 

The network architecture includes three convolution layers, two general fully connected layers and a $N$-branch fully connected layer, where $N$ is the number of training sequences. To distinguish the foreground and background object in the tracking procedure, each of the last branches includes a binary softmax classifier with cross-entropy loss. As compared with other CNN-based trackers, the MDNet has fewer layers, which lowers the complexity of an online testing procedure and has a more precise localization prediction. During online tracking, the $N$-branch fully connected layer is replaced by a single-branch layer. Besides, the weights in the first five layers are pretrained during the multi-domain learning, and random initialization is exploited to the weights in the new single-branch layer. The weights in the fully connected layer are updated during online tracking whereas the weights in the convolutional layers are frozen. Both the general and domain-specific features are preserved in this strategy and the tracking speed is improved as well. 

To control the tracking procedure and weights update, long-term and short-term updates are conducted respectively based on length of the consistent positive examples intervals. Besides, hard negative example mining \cite{shrivastava2016training} is performed to reduce the positive/negative example ratio and improve the binary classification difficulty to make the network more discriminative. Finally, bounding box regression is exploited to adjust the accurate target location.

\begin{figure}[ht]
\centering 

\subfloat{\includegraphics[width=0.32\linewidth]{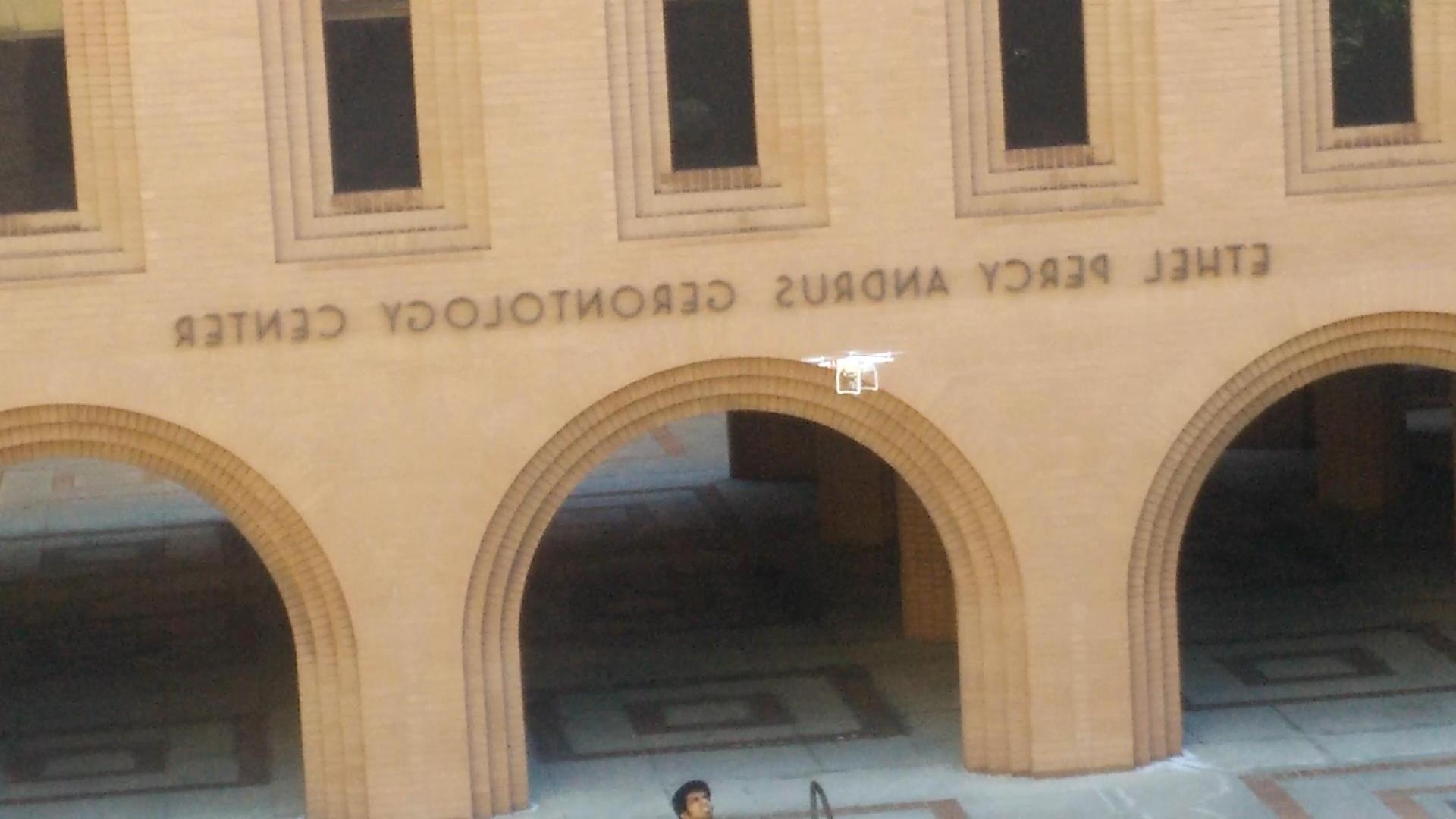}}  \hfil 
\subfloat{\includegraphics[width=0.32\linewidth]{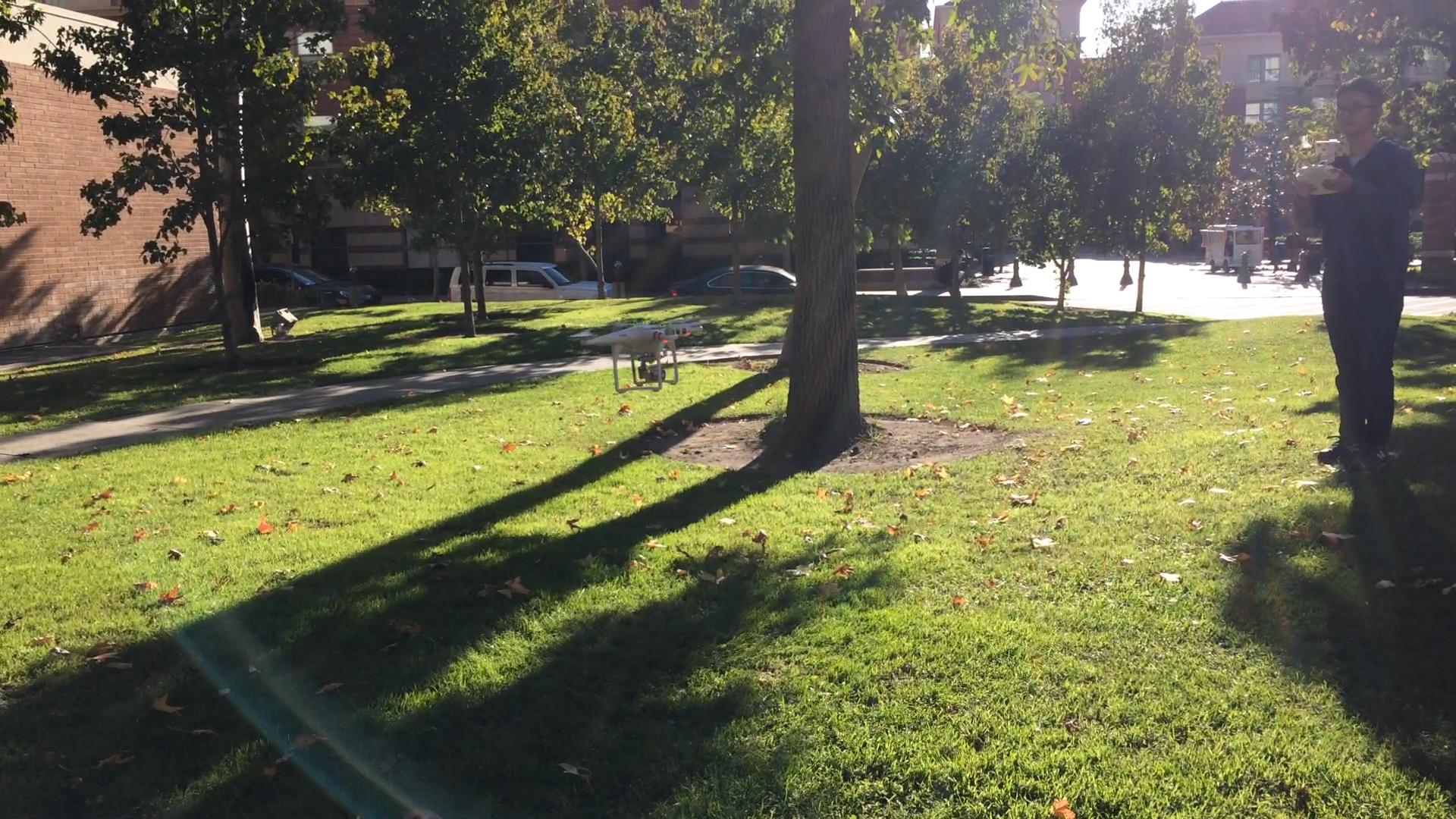}}  \hfil 
\subfloat{\includegraphics[width=0.32\linewidth]{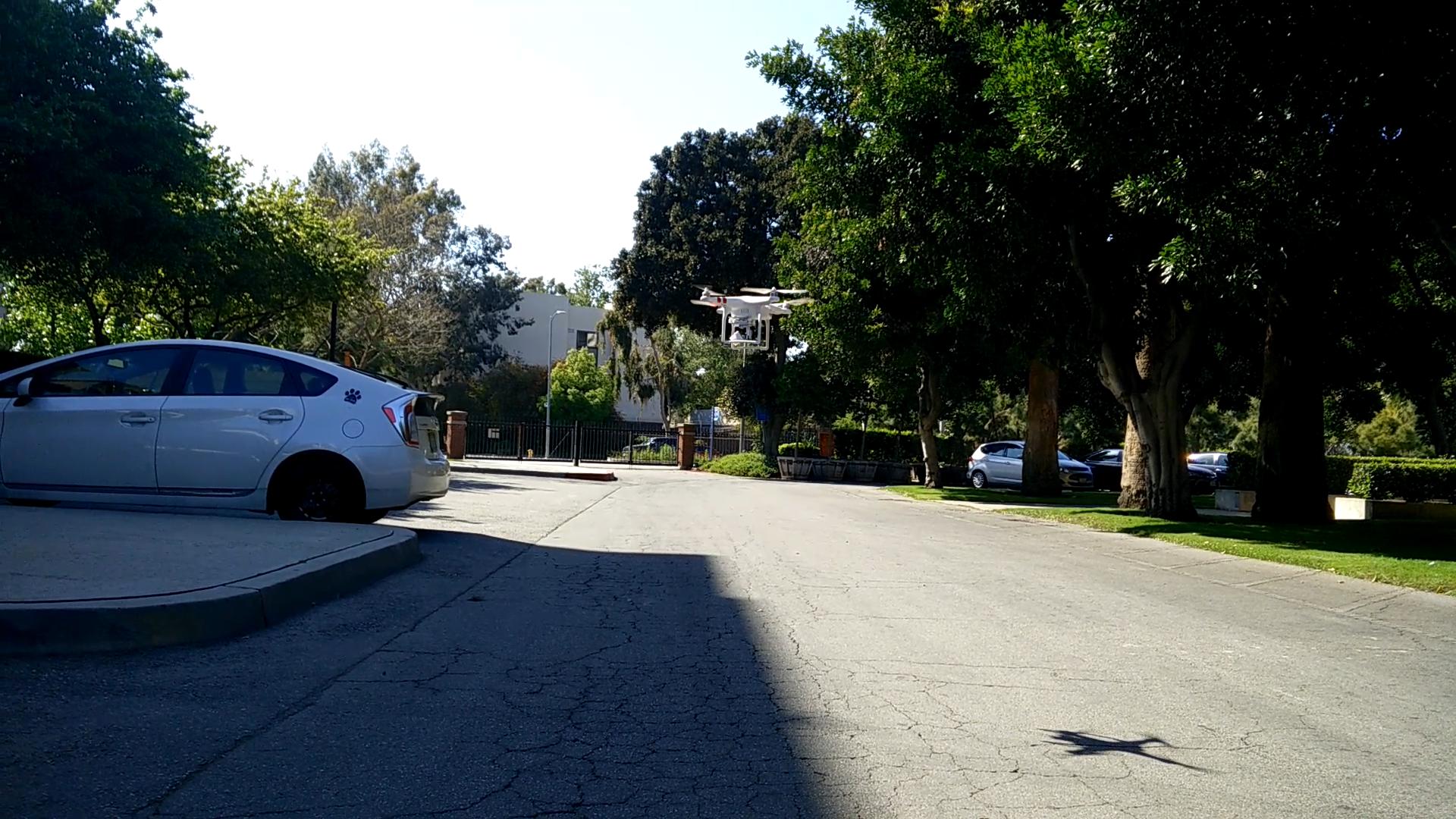}}  \hfil  \\
% \subfloat{\includegraphics[width=0.24\linewidth]{fig/compare/detection/img1939.png}}  \hfil  \\

\vspace{-0.10in}

\subfloat{\includegraphics[width=0.32\linewidth]{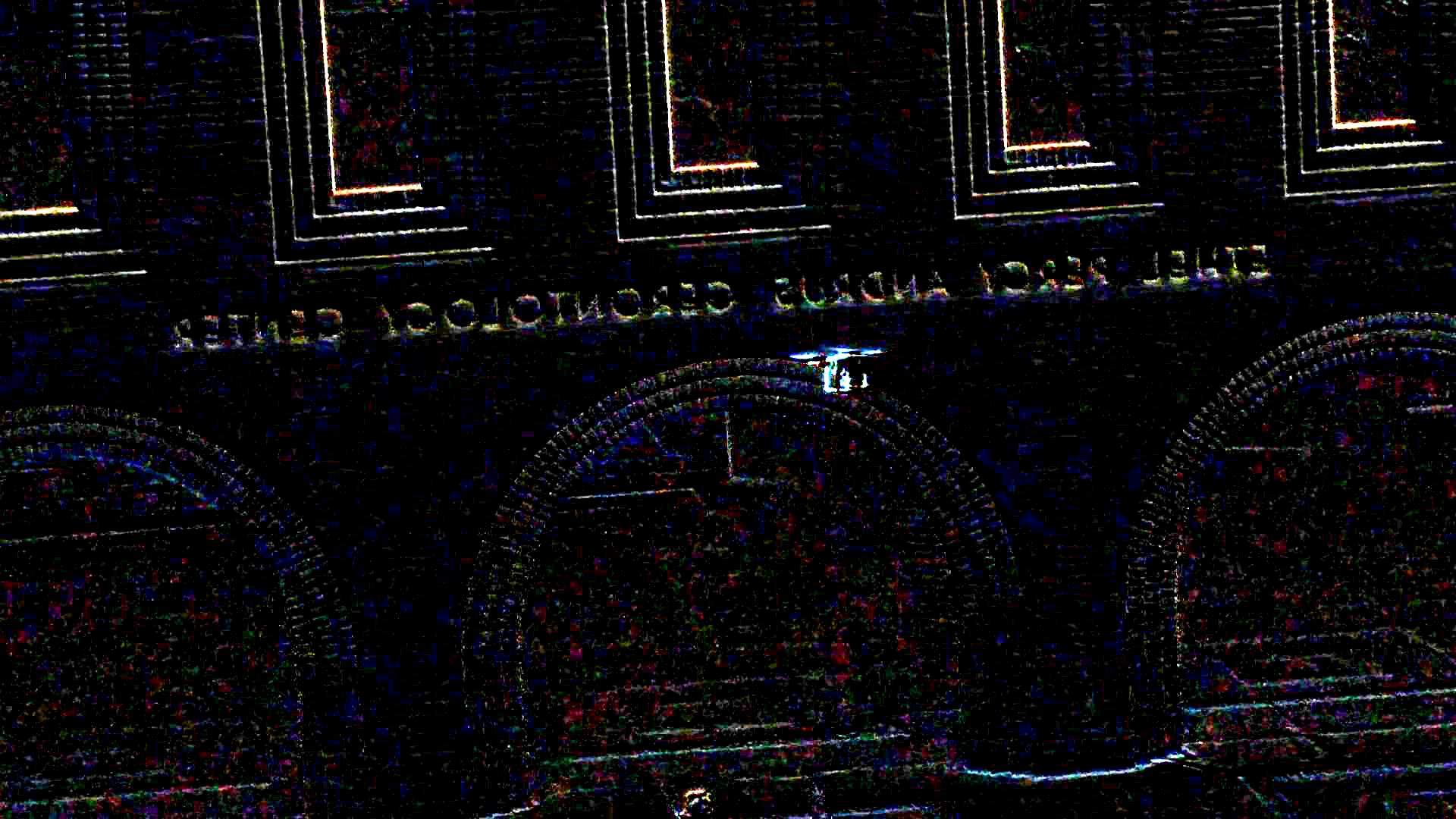}}  \hfil 
\subfloat{\includegraphics[width=0.32\linewidth]{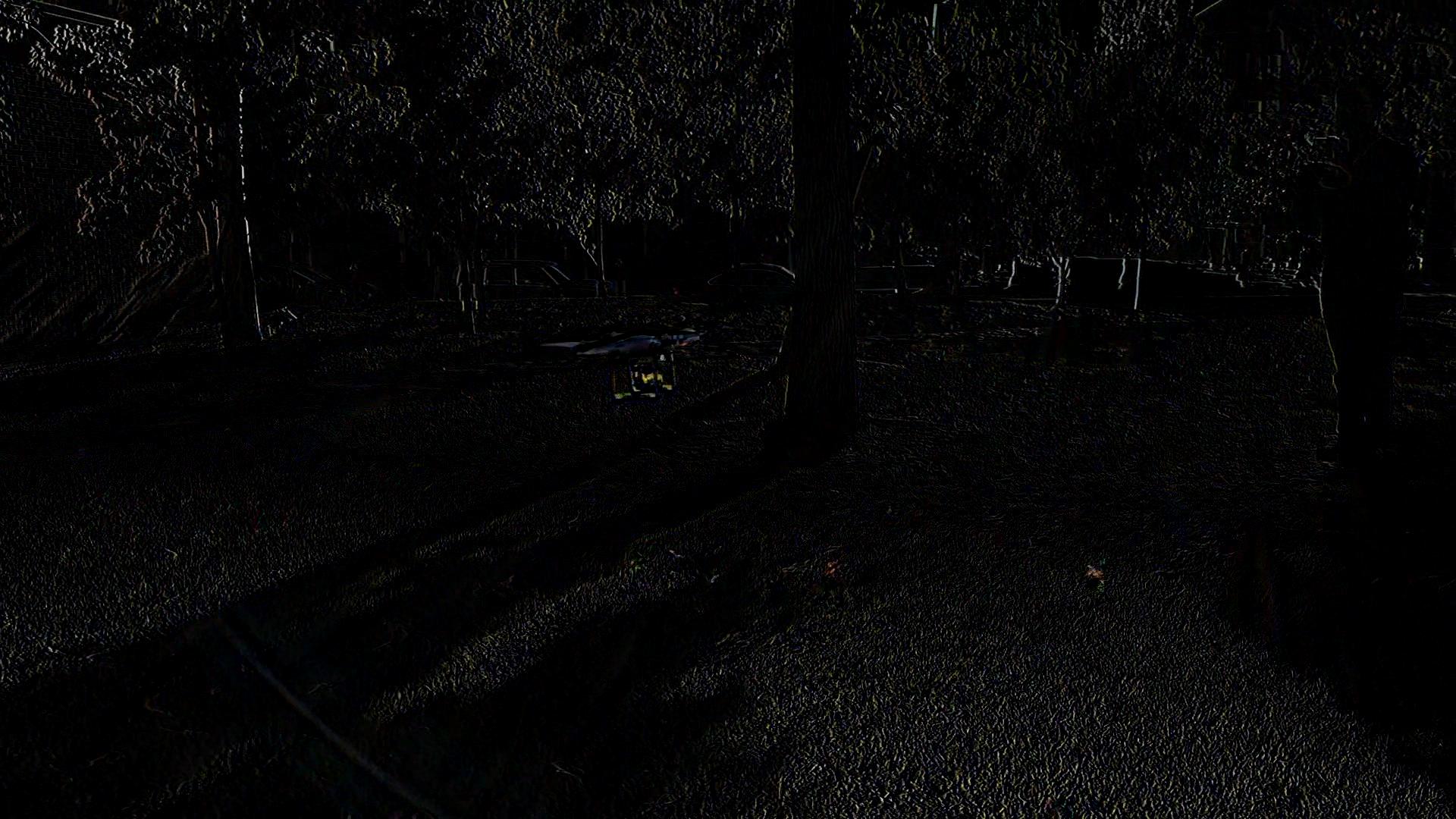}}  \hfil 
\subfloat{\includegraphics[width=0.32\linewidth]{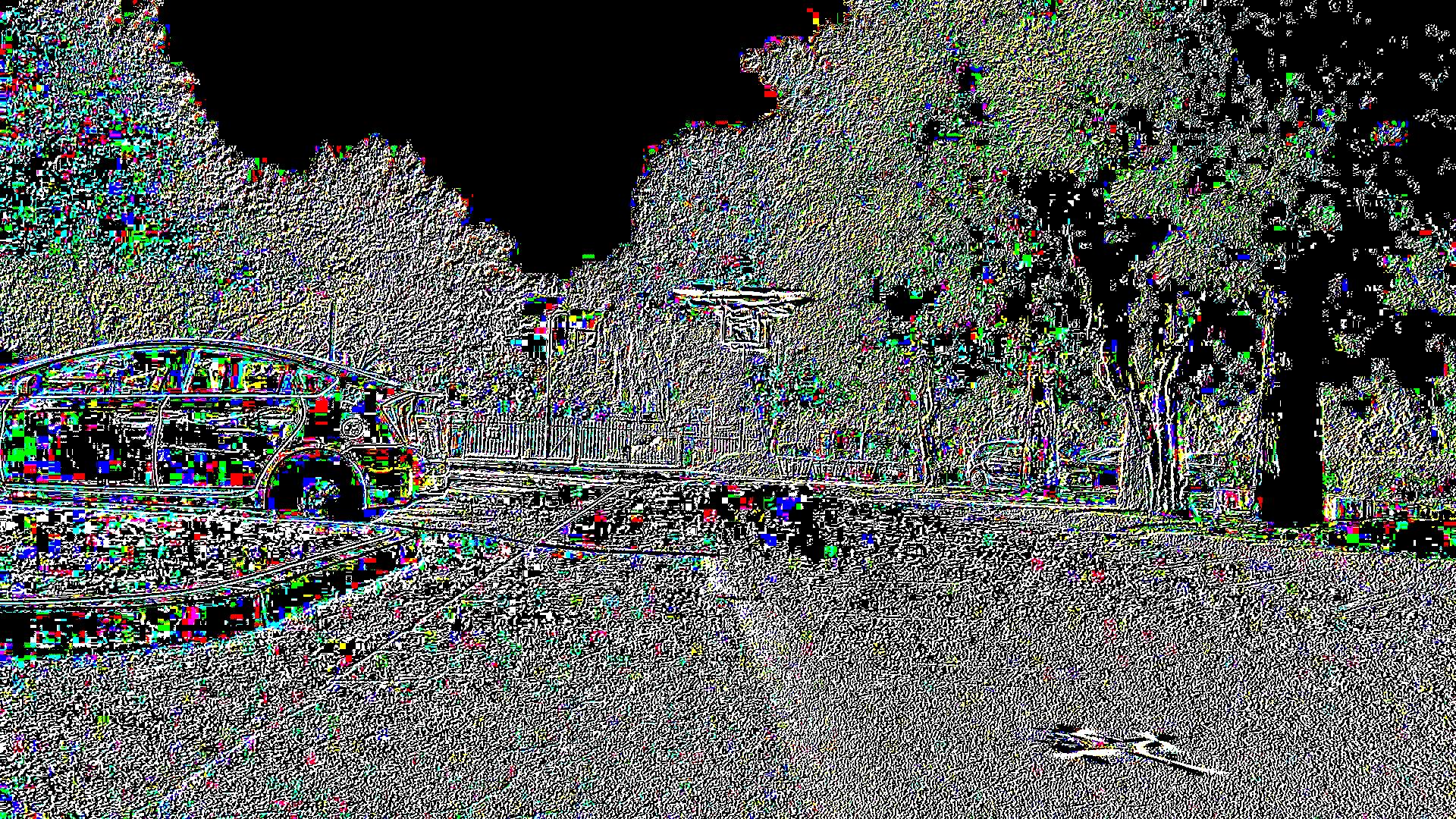}}  \hfil  \\

\caption{Comparison of three raw input images (first row) and their corresponding
residual images (second row).}\label{fig:residueimage}
\end{figure}

To improve the tracking performance furthermore, we propose a video
pre-processing step. That is, we subtract the current frame from the
previous frame and take the absolute values pixelwise to obtain the
residual image of the current frame.  Note that we do the same for the
R,G,B three channels of a color image frame to get a color residual
image.  Three color image frames and their corresponding color residual
images are shown in Fig.  \ref{fig:residueimage} for comparison.  If
there is a panning movement of the camera, we need to compensate the
global motion of the whole frame before the frame subtraction operation.

%%%%%%%%%%%%%%%%%%%%%%%%%%%%%%%%%%%%%

%%%%%%%%%%%%%%%%%%%%%%%%%%%%%%%%%%%%%

Since there exists strong correlation between two consecutive images,
most background of raw images will cancel out and only the fast moving
object will remain in residual images. This is especially true when the
drone is at a distance from the camera and its size is relatively small.
The observed movement can be well approximated by a rigid body motion.
We feed the residual sequences to the MDNet for drone tracking after the
above pre-processing step. It does help the MDNet to track the drone
more accurately. Furthermore, if the tracker loses the drone for a short
while, there is still a good probability for the tracker to pick up the
drone in a faster rate. This is because the tracker does not get
distracted by other static objects that may have their shape and color
similar to a drone in residual images. Those objects do not appear in
residual images.

\subsection{Integrated Detection and Tracking System}\label{sec:fusion}

\begin{figure}[ht]
\begin{center}
\includegraphics[width=0.95\linewidth]{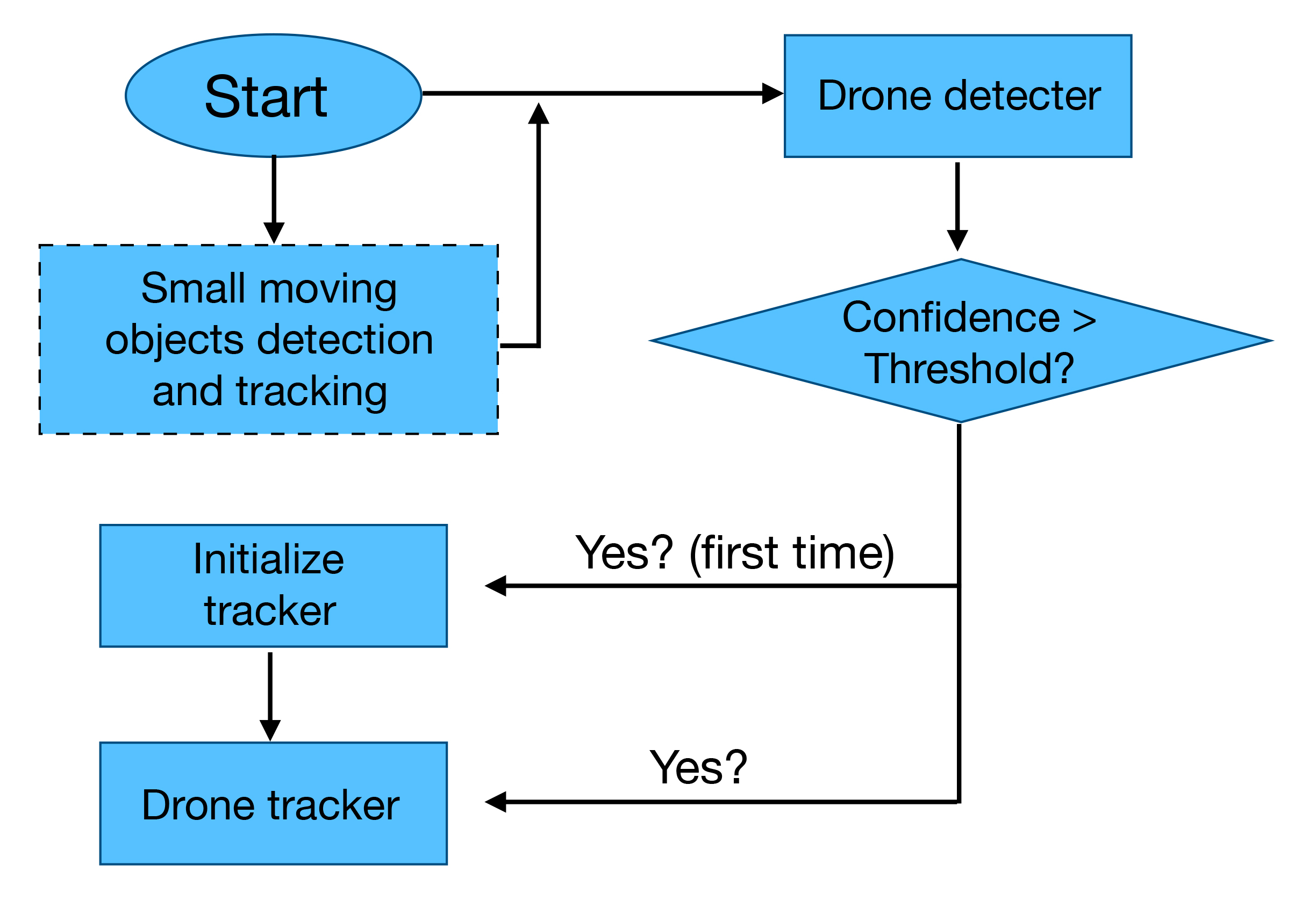}
\end{center}
\caption{A flow chart of the drone monitoring system.}\label{fig:overview}
\end{figure}

There are limitations in detection-only or tracking-only modules.  The
detection-only module does not exploit the temporal information, leading
to huge computational waste. The tracking-only module attempts to track the drone by leveraging the temporal relationships between video frames without knowing the object information, but it cannot initialize the drone tracker when failed to track for a certain time interval. To build a complete system, we need to integrate these two modules into one.  The
flow chart of the proposed drone monitoring system is shown in Fig.
\ref{fig:overview}. 

%%%%%%%%%%%%%%%%%%%%%%%%%%%%%%%%%%%%%
%%%%%%%%%%%%%%%%%%%%%%%%%%%%%%%%%%%%%

%%%%%%%%%%%%%%%%%%%%%%%%%%%%%%%%%%%%%
\begin{figure}[t!]
\begin{center}
    \begin{subfloat}[Synthetic Dataset]{ 
        \centering
        \includegraphics[width=0.95\linewidth]{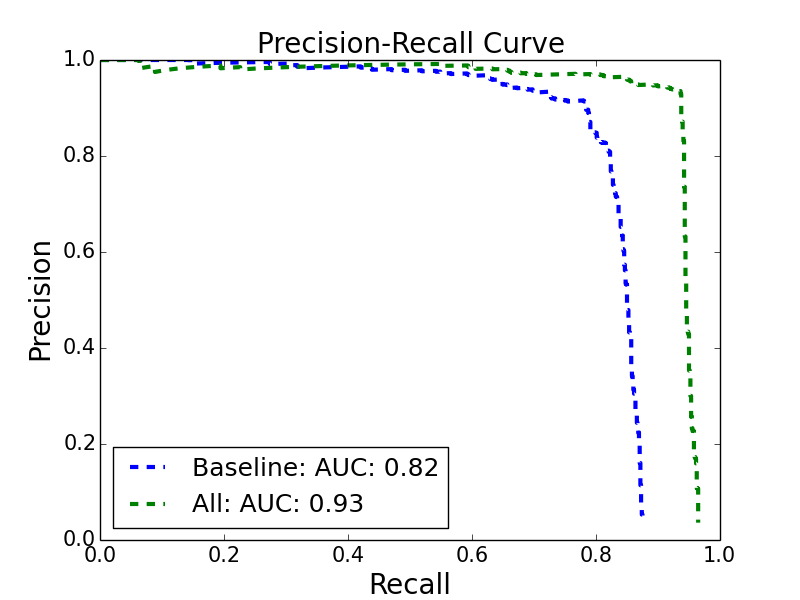}\label{fig:detectorRa}}
        % \caption{Synthetic Dataset} \label{fig:detectorRa}
    \end{subfloat}
    \begin{subfloat}[Real-World Dataset]{
        \centering
        \includegraphics[width=0.95\linewidth]{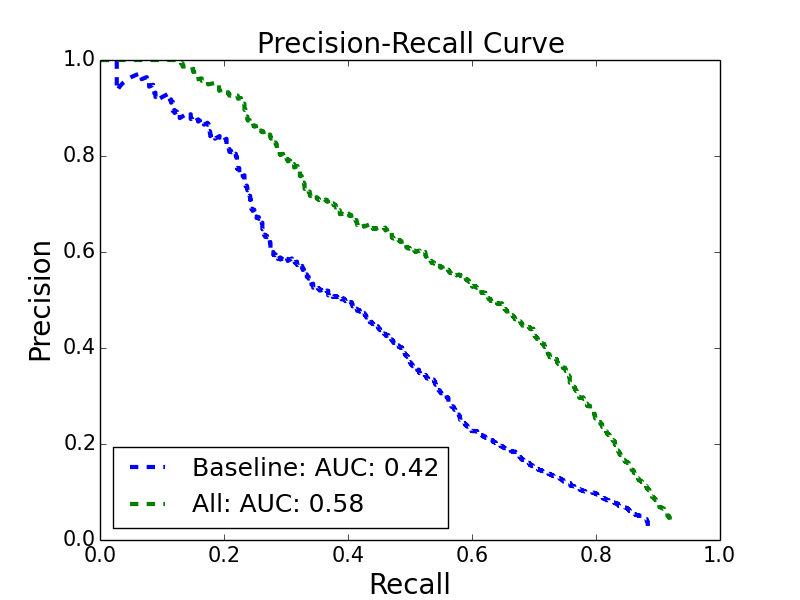}\label{fig:detectorRb}}
        % \caption{Real-World Dataset} \label{fig:detectorRb}
    \end{subfloat}
\end{center}
\caption{Comparison of the visible drone detection performance on (a) the
synthetic and (b) the real-world datasets, where the baseline method
refers to that using geometric transformations to generate training data
only while the All method indicates that exploiting geometric transformations,
illumination conditions and image quality simulation for data
augmentation.} \label{fig:detectorRe}
\end{figure} 
%%%%%%%%%%%%%%%%%%%%%%%%%%%%%%%%%%%%%

%%%%%%%%%%%%%%%%%%%%%%%%%%%%%%%%%%%%%
\begin{figure}[t!]
\begin{center}
    \begin{subfloat}[Synthetic Dataset]{ 
        \centering
        \includegraphics[width=0.95\linewidth]{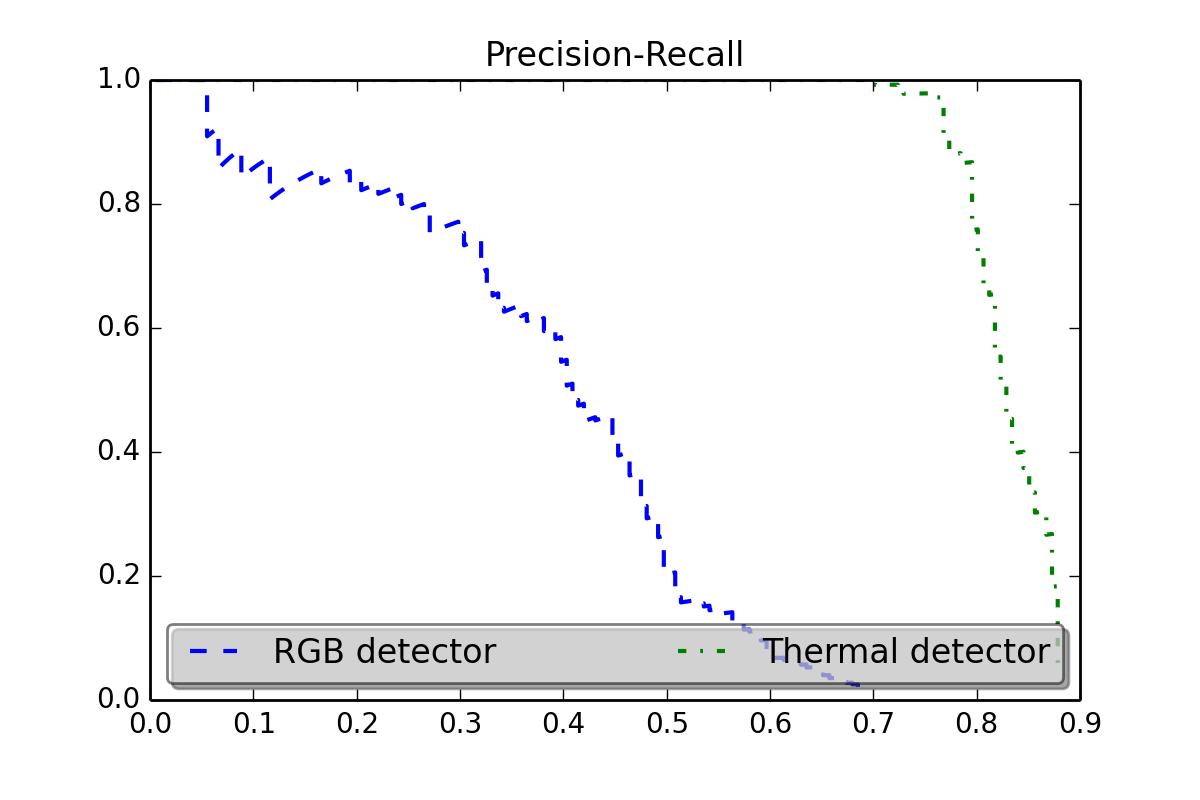}\label{fig:detectort1}}
    \end{subfloat}
    \begin{subfloat}[Real-World Dataset]{
        \centering
        \includegraphics[width=0.95\linewidth]{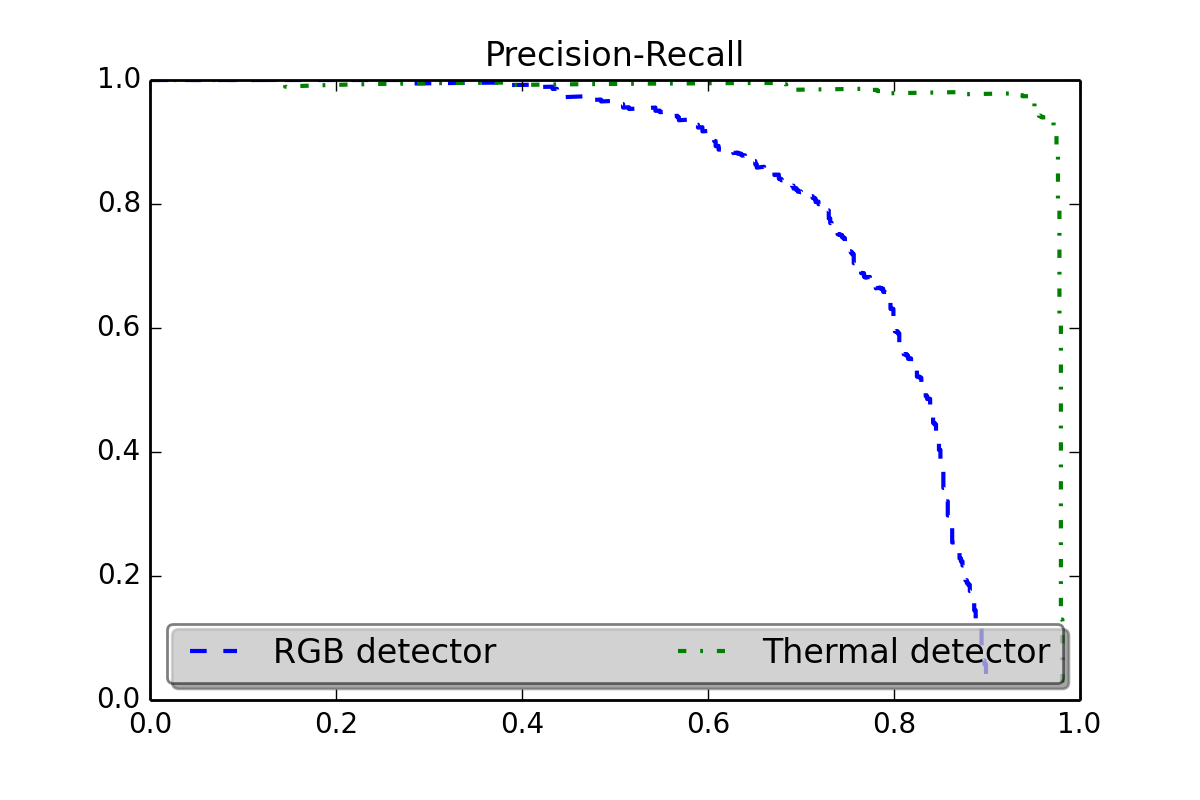}\label{fig:detectort2}}
    \end{subfloat}
    \begin{subfloat}[Real-World Dataset]{
        \centering
        \includegraphics[width=0.95\linewidth]{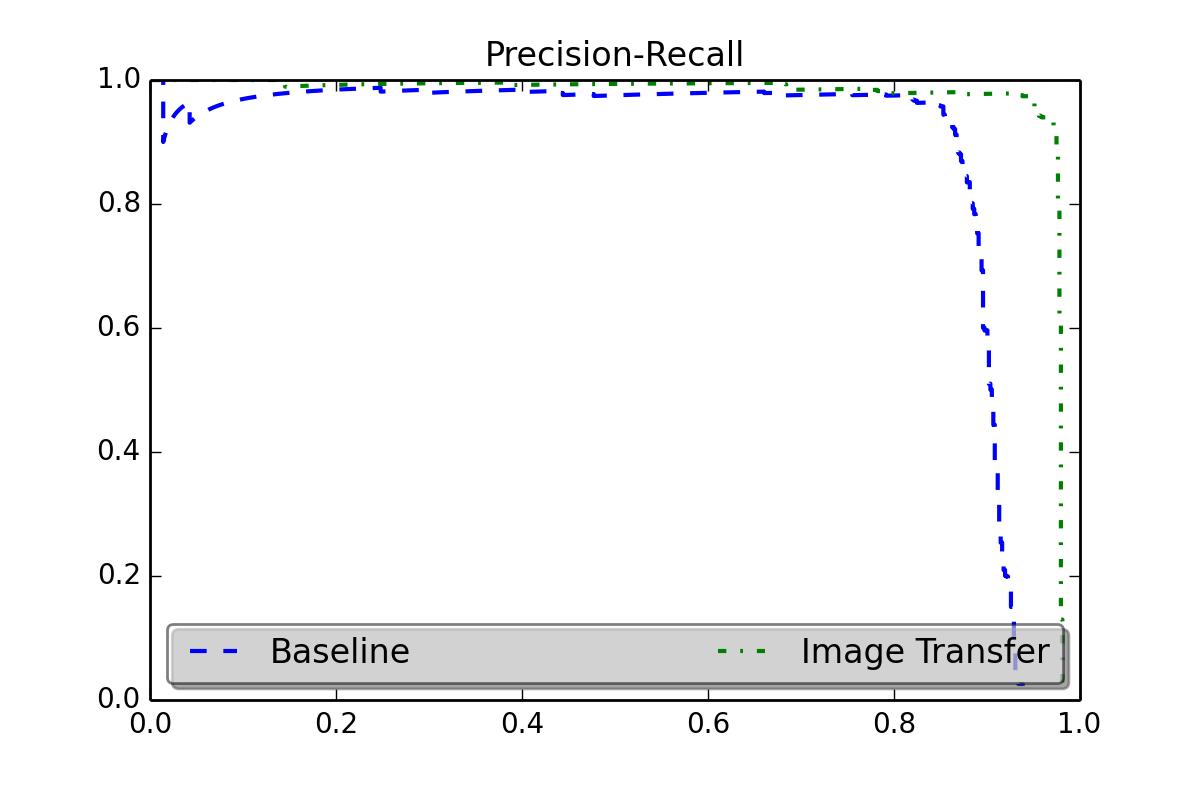}\label{fig:detectort3}}
    \end{subfloat}
\end{center}
\caption{Comparison of the visible and thermal drone detection performance on (a) the
synthetic, (b) the real-world datasets. (c) shows the comparison of different thermal data augmentation techniques, where the baseline method
refers to that using geometric transformations, illumination conditions and image quality simulation for data augmentation, and image transfer refers to that utilizing the proposed modified Cycle-GAN data augmentation technique.} \label{fig:detectorR}
\end{figure} 
%%%%%%%%%%%%%%%%%%%%%%%%%%%%%%%%%%%%% 

Generally speaking, the drone detector has two tasks -- finding the
drone and initializing the tracker. Typically, the drone tracker is used
to track the detected drone after the initialization.  However, the
drone tracker can also play the role of a detector when an object is too
far away to be robustly detected as a drone due to its small size. Then,
we can use the tracker to track the object before detection based on the
residual images as the input. Once the object is near, we can use the
drone detector to confirm whether it is a drone or not. 

An illegal drone can be detected once it is within the field of view and
of a reasonable size. The detector will report the drone location to the
tracker as the start position. Then, the tracker starts to work.  During
the tracking process, the detector keeps providing the confidence score
of a drone at the tracked location as a reference to the tracker.  The
final updated location can be acquired by fusing the confidence scores
of the tracking and the detection modules as follows. 

For a candidate bounding box, we can compute the confidence scores of
this location via
\begin{eqnarray} \label{eqn:confidence}
    S'_d&=& 1 / ({1+e^{-\beta_1(S_d-\alpha_1)}}),\\
    S'_t&=& 1 / ({1+e^{-\beta_2(S_t-\alpha_2)}}),\\
    S' &=& \max(S'_d, S'_t),
\end{eqnarray}  
where $S_d$ and $S_t$ denote the confidence scores obtained by the
detector and the tracker, respectively, $S'$ is the confidence score
of this candidate location and parameters $\beta_1$, $\beta_2$,
$\alpha_1$, $\alpha_2$ are used to control the acceptance threshold. 

We compute the confidence score of a couple of bounding box candidates, where $S'_i$ denotes the confidence score and $BB_i$ denotes the bounding box position, $i \in C$, where $C$ denotes the set of candidate
indices. Then, we select the one with the highest score:
\begin{eqnarray}
i^* & = & \underset{i \in C}{\operatorname{argmax}}~S'_i, \\
S_f & = & \underset{i \in C}{\operatorname{max}}~S'_i, \\
BB_{i^*} & = &  BB_{\underset{i \in C}{\operatorname{argmax}}~S'_i}
\end{eqnarray}  
where $BB_{i^*}$ is the finally selected bounding box and $S_f$ is its
confidence score. If $S_f = 0$, the system will report a message of
rejection. 

\section{Experimental Results}\label{sec:results}

\subsection{Drone Detection}

We test the visible and thermal drone detector on both the real-world and the synthetic visible or thermal datasets. Each of them contains 1000 images. The images in the real-world dataset are sampled from
videos in the USC Drone dataset and the USC Thermal Drone dataset. The images in the synthetic dataset are generated using different foreground and background images with those in the training dataset. The detector can take any size of images as the input.
These images are then re-scaled such that their shorter side has 600
pixels \cite{fasterrcnn}. To evaluate the drone detector, we compute the precision-recall curve.
Precision is the fraction of the total number of detections that are
true positive. Recall is the fraction of the total number of labeled
samples in positive class that are true positive. The area under the
precision-recall curve (AUC) \cite{auc} is also reported.  

As for the visible detector, the effectiveness of the proposed data augmentation technique is illustrated in Fig. \ref{fig:detectorRe}. In this figure, we compare the performance
of the baseline method that uses simple geometric transformations only
and that of the method that uses all mentioned data augmented
techniques, including geometric transformations, illumination conditions
and image quality simulation. Clearly, better detection performance can
be achieved by more augmented data.  We see around $11\%$ and $16\%$
improvements in the AUC measure on the real-world and the synthetic
datasets, respectively. 

The experiments results of thermal drone detector are presented in Fig. \ref{fig:detectorR}. In both real-world and synthetic thermal datasets, the thermal detector achieves better performance than the visible detector by a large margin. We further compare the proposed modified Cycle-GAN data augmentation approach with the traditional data augmentation techniques to present the necessity of the proposed approach. The comparison between baseline methods which uses the 3D rendering techniques and image transfer methods which exploits the modified Cycle-GAN model for generating foreground images are demonstrated in Fig. \ref{fig:detectorR}. We observe there is $8\%$ performance gain in the AUC measure on the real-world datasets.

\subsection{Drone Tracking}

The MDNet is adopted as the object tracker.  We take 3 video sequences
from the USC drone dataset as testing ones. They cover several
challenges, including scale variation, out-of-view, similar objects in
background, and fast motion.  Each video sequence has a duration of 30
to 40 seconds with 30 frames per second. Thus, each sequence contains
900 to 1200 frames. Since all video sequences in the USC drone dataset
have relatively slow camera motion, we can also evaluate the advantages
of feeding residual frames (instead of raw images) to the MDNet. 

The performance of the tracker is measured with the area-under-the-curve
(AUC) measure.  We first measure the intersection over union $(IoU)$ for all
frames in all video sequences as
\begin{equation}
IoU = \frac{Area~ of~ Overlap}{Area~ of~ Union}, 
\end{equation}  
where the ``Area of Overlap" is the common area covered by the predicted
and the ground truth bounding boxes and the ``Area of Union" is the
union of the predicted and the ground truth bounding boxes.  The IoU
value is computed at each frame. If it is higher than a threshold, the
success rate is set to 1; otherwise, 0.  Thus, the success rate value is
either 1 or 0 for a given frame.  Once we have the success rate values
for all frames in all video sequences for a particular threshold, we can
divide the total success rate by the total frame number. Then, we can
obtain a success rate curve as a function of the threshold. Finally, we
measure the area under the curve (AUC) which gives the desired
performance measure. 

\begin{figure}[t!]
\begin{center}
\includegraphics[width=0.95\linewidth]{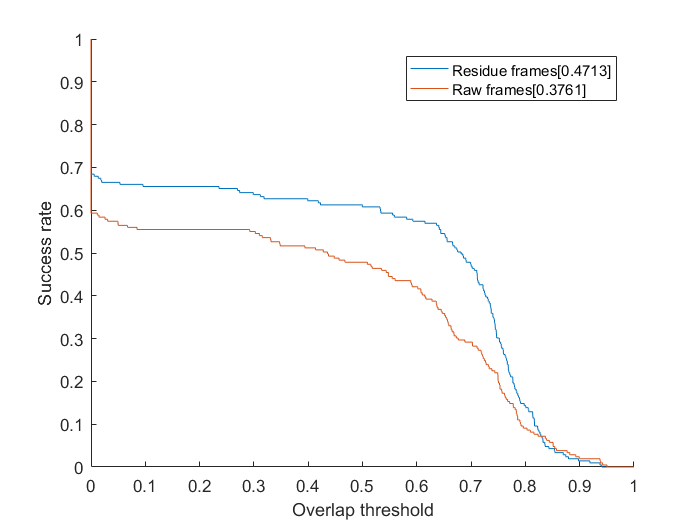}
\end{center}
\caption{Comparison of the MDNet tracking performance using the raw 
and the residual frames as the input.} \label{fig:trackR}
\end{figure}

We compare the success rate curves of the MDNet using the original
images and the residual images in Fig. \ref{fig:trackR}.  As compared to
the raw frames, the AUC value increases by around 10\% using the
residual frames as the input. It collaborates the intuition that
removing background from frames helps the tracker identify the drones
more accurately. Although residual frames help improve the performance
of the tracker for certain conditions, it still fails to give good
results in two scenarios: 1) movement with fast changing directions and
2) co-existence of many moving objects near the target drone.  To
overcome these challenges, we have the drone detector operating in
parallel with the drone tracker to get more robust results. 

\begin{figure*}[ht]
\begin{center}
\includegraphics[width=0.95\linewidth]{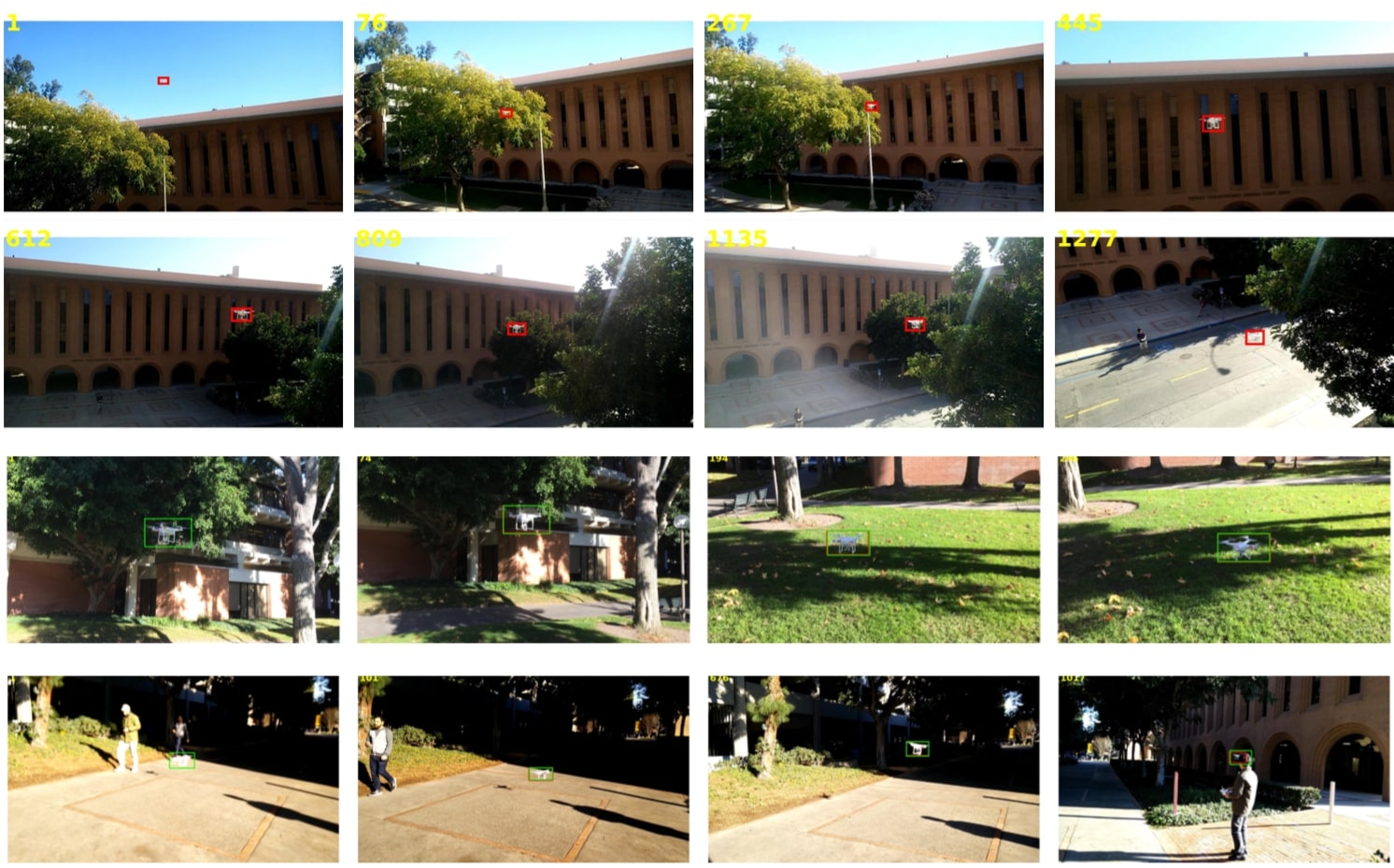}
\end{center}
\caption{Qualitative results on USC drone datasets. Our algorithm performs well on small object tracking and long sequence (first and second row), complex background (third row), and occlusion (fourth row). The bounding boxes in red are integrated system results and the bounding boxes in green are tracking-only results.}\label{fig:drone_tracking}
\end{figure*}

Our algorithm performs well for most of the sequences on USC drone datasets as shown in Fig. \ref{fig:drone_tracking}. The qualitative results show that our algorithm performs well for tracking drones in distances and a long video (first and second row). The third row shows that both tracking algorithm (in green) and integrated system (in red) works well when meeting with complex background. The fourth row shows our approach performs well with motion blur and occlusions. Especially the integrated system predicts more accurate and tight bounding boxes in red compared with those of the tracking-only algorithm in green in the fourth column.

%%%%%%%%%%%%%%%%%%%%%%%%%%%%%%%%%%%%%{}{}
%%%%%%%%%%%%%%%%%%%%%%%%%%%%%%%%%%%%%

%%%%%%%%%%%%%%%%%%%%%%%%%%%%%%%%%%%%%
\begin{figure*}[ht]
\begin{center}
\includegraphics[width=0.95\linewidth]{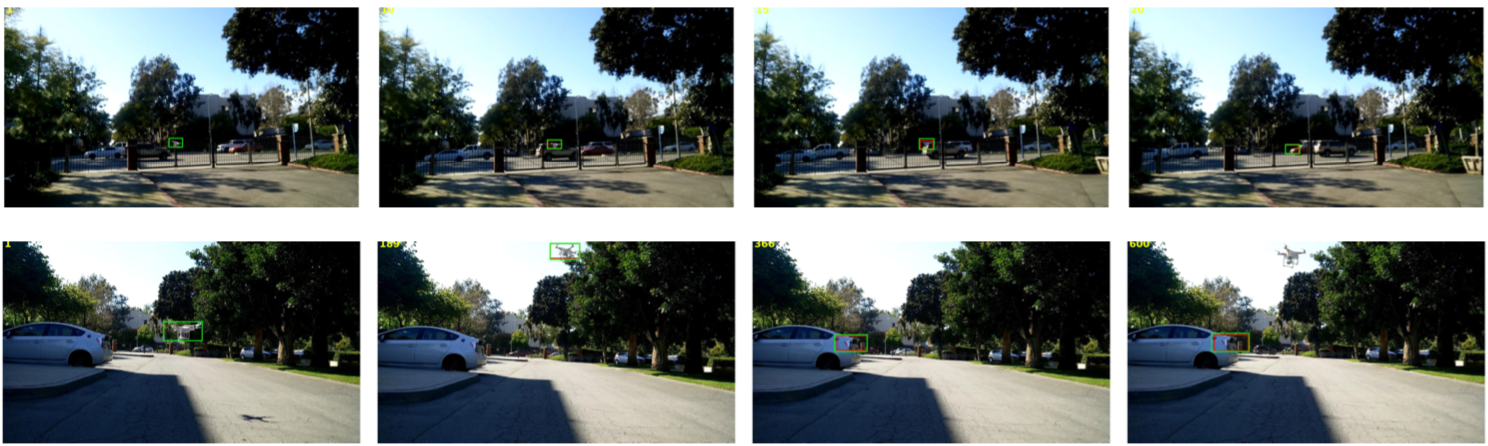}
\end{center}
\caption{Failure cases. Our algorithm fails to track the drone with strong motion blur and complex background (top row), and fails to re-identify the drone when it goes out-of-view and back (bottom row). The bounding boxes in red are integrated system results and the bounding boxes in green are tracking-only results.}\label{fig:failure}
\end{figure*}
%%%%%%%%%%%%%%%%%%%%%%%%%%%%%%%%%%%%%

We present failure cases on USC drone datasets in Fig. \ref{fig:failure}. The first row shows both tracking algorithm and integrated system fail when dealing with strong motion blur, complex background in the fourth column. The second row shows the failure case when the drone is out-of-view for a long time where the drone is going out of view in the frame 189 and fully comes back in the frame 600. Both the two approaches cannot track the drone when the drone reappears in the video. Long term memory should be applied in this case and optical flow could be exploited to re-identify the drone to initialize the tracker. 

%%%%%%%%%%%%%%%%%%%%%%%%%%%%%%%%%%%%%
%%%%%%%%%%%%%%%%%%%%%%%%%%%%%%%%%%%%%

\subsection{Fully Integrated System}

%%%%%%%%%%%%%%%%%%%%%%%%%%%%%%%%%%%%%
\begin{figure}[ht]
\begin{center}
\includegraphics[width=0.95\linewidth]{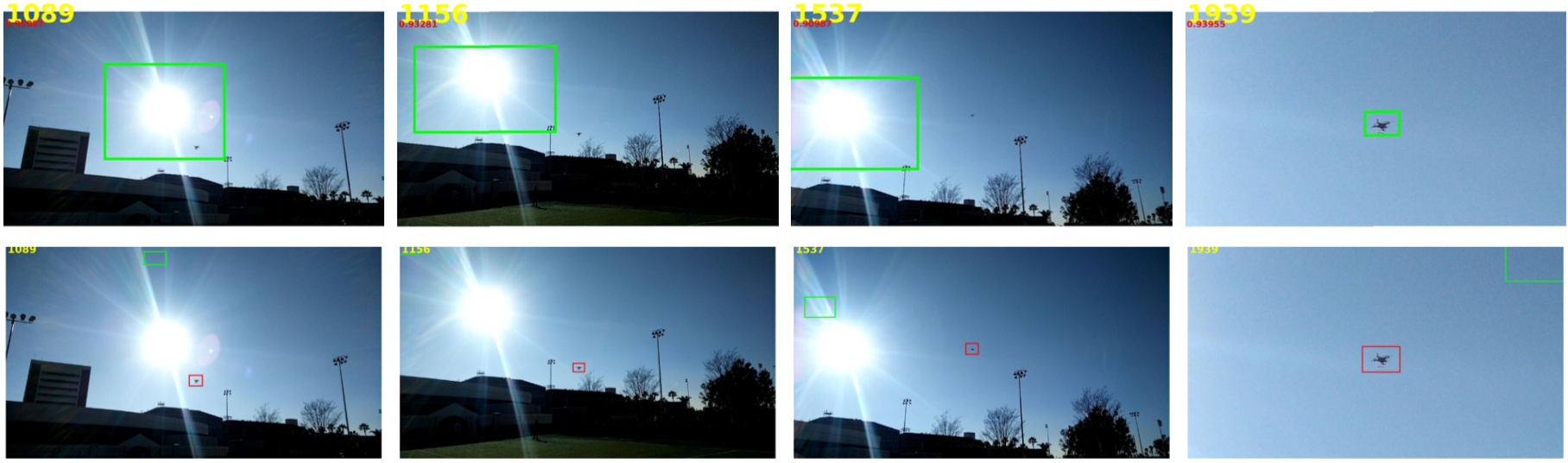}
\end{center}

\caption{Comparison of qualitative results of detection, tracking and integrated system on the $drone\_garden$ sequence. The detection results are shown in the first row. The corresponding tracking and integrated system results are shown in the second row with tracking bounding boxes in green and integrated system bounding boxes in red respectively.}\label{fig:compare}
\end{figure}
%%%%%%%%%%%%%%%%%%%%%%%%%%%%%%%%%%%%%  

%%%%%%%%%%%%%%%%%%%%%%%%%%%%%%%%%%%%%
\begin{figure}[ht]
\begin{center}
\includegraphics[width=0.95\linewidth]{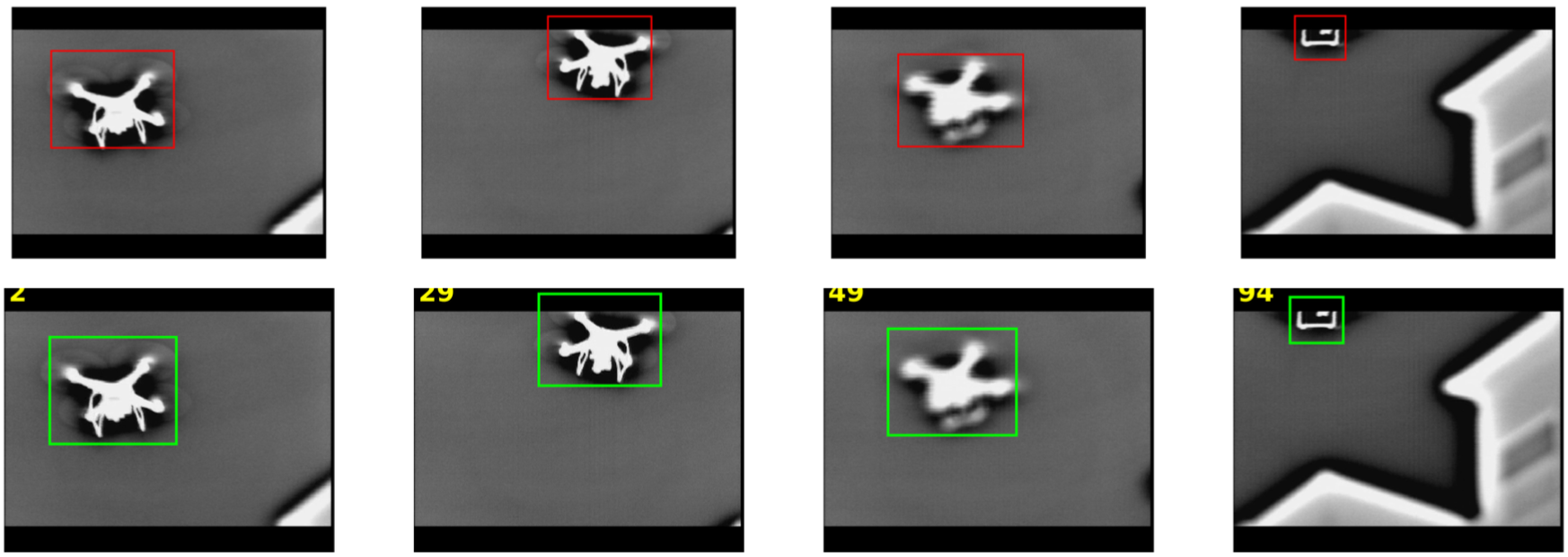}
\end{center}

\caption{Comparison of qualitative results of detection (first row), integrated system (second row) on the thermal drone sequence.}\label{fig:thermal_d&t}
\end{figure}
%%%%%%%%%%%%%%%%%%%%%%%%%%%%%%%%%%%%%

\begin{figure}[t!]
\begin{center}
       \includegraphics[width=0.95\linewidth]{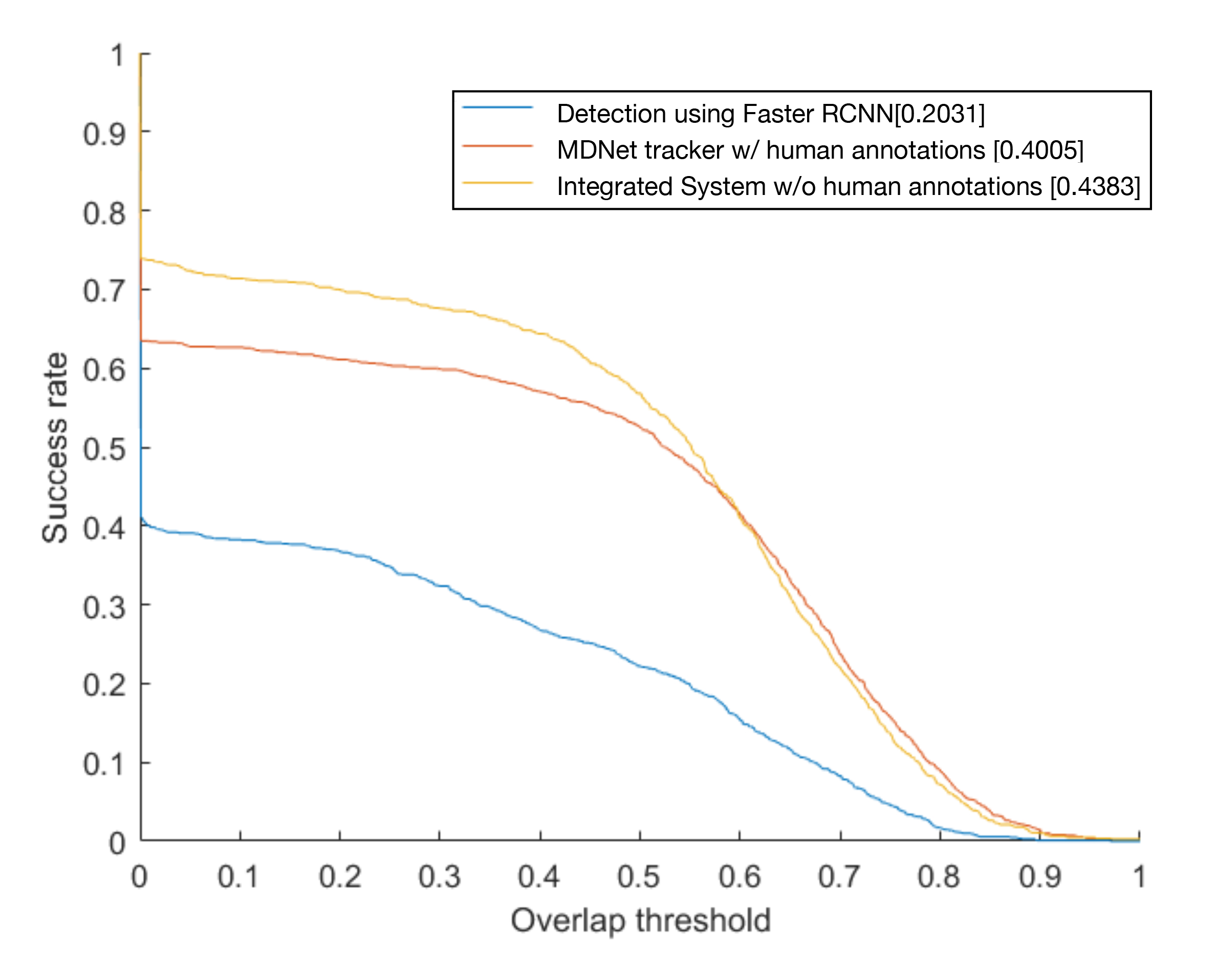}
\end{center}
\caption{Detection only (Faster RCNN) vs. tracking only (MDNet tracker)
vs. our integrated system: The performance increases when we fuse the
detection and tracking results.} \label{fig:systemR}
\end{figure}

The fully integrated system contains both the detection and the tracking
modules. We use the USC drone dataset to evaluate the performance of the
fully integrated system.  The performance comparison (in terms of the
AUC measure) of the fully integrated system, the conventional MDNet (the
tracker-only module) and the Faster-RCNN (the detector-only module) is
shown in Fig. \ref{fig:systemR}. Note that the tracker-only module needs human annotation for the first frame to perform the drone tracking, while the integrated system is an autonomous system without relying on the first frame annotation. The fully integrated system
outperforms the other benchmarking methods by substantial margins. This
is because the fully integrated system can use detection as the means to
re-initialize its tracking bounding box when it loses the object. It is worth pointing out that the overall detection performance is not good, moreover the ground truth label is more accurate than the detection results which leads to that tracking has better success rate when the overlap threshold is larger than 0.8.

We show the comparison of qualitative results of detection, tracking and integrated system in Fig. \ref{fig:compare}. The drone detection-only results in the first row demonstrates that detection-only performs bad when there are strong illumination changes. The drone tracker in green in the second row failes to track the drones back when it loses tracking at the very beginning. The proposed integrated system has higher tolerance against the illuminance and scale changes which are in the red bounding boxes in the second row. It outperforms the detection-only resuls and tracking-only results since it learns from both of them. Furthermore, we present the comparison of qualitative results of the thermal detection and integrated system in Fig. \ref{fig:thermal_d&t}. Both the two approaches perform well to monitor the drone at night. Since the thermal drone detector has over $8\%$ performance gain in the AUC measure over the visible drone detector and it almost achieves ``perfect'' precision-recall curve. We apply the object detection algorithm every other $n$ frames, and re-initialize the tracker if the detection confidence score is higher than the threshold. Due to the high detection performance, the tracklet is updated nearly every $n$ frames by the detection. Therefore, the integrated system has similar performance with detection results in this case.

%%%%%%%%%%%%%%%%%%%%%%%%%%%%%%%%%%%%%
%%%%%%%%%%%%%%%%%%%%%%%%%%%%%%%%%%%%%

\section{Conclusion}\label{sec:conclusion}

A video-based drone monitoring system was proposed in this work to detect and track drones during day and night.  The
system consists of the drone detection module and the drone tracking
module. Both of them were designed based on deep learning networks.  We
developed a model-based data augmentation technique for visible drone monitoring to enrich the
training data. Besides, we presented an adversarial data augmentation methodology to create more thermal drone images due to the lack of the thermal drone data. We also exploited residue images as the input to the
drone tracking module. The fully integrated monitoring system takes
advantage of both modules to achieve high performance monitoring.
Extensive experiments were conducted to demonstrate the superior
performance of the proposed drone monitoring system. 

\section*{Acknowledgment}

This research is supported by a grant from the Pratt \& Whitney
Institute of Collaborative Engineering (PWICE).

\bibliographystyle{achemso}
\bibliography{egbib}

\end{document}